\pdfoutput=1
\documentclass[final,3p,times]{elsarticle}

\usepackage{amssymb}
\usepackage{amsmath}
\usepackage{bm}
\usepackage{graphicx}
\usepackage{subcaption}
\usepackage{caption}
\usepackage{booktabs}
\usepackage{multirow}
\usepackage{makecell}
\usepackage{placeins}
\usepackage{algorithm}
\usepackage{algpseudocode}
\usepackage[markup=default]{changes}
\usepackage{xurl}

\providecommand{\SubmissionJournal}{Computer-Aided Design}
\journal{\SubmissionJournal}

\begin{document}

\begin{frontmatter}

\author[zju]{Ziheng Dai} %
\author[hdu]{Wanbin Pan}
\author[zju]{Zhihong Tang}
\author[zju]{Shuming Gao\corref{cor1}}
\cortext[cor1]{Corresponding author \\ \textit{E-mail address}: smgao@cad.zju.edu.cn}
\affiliation[zju]{organization={State Key Laboratory of CAD\&CG},%
            addressline={Zhejiang University},
            city={Hangzhou},
            postcode={310058},
            state={Zhejiang},
            country={People’s Republic of China}}
\affiliation[hdu]{organization={School of Media and Design},%
            addressline={Hangzhou Dianzi University},
            city={Hangzhou},
            postcode={310001},
            state={Zhejiang},
            country={People’s Republic of China}}
\title{Automatic assembly modeling through attention-based assembly feature mate prediction and assembly feature pattern recognition}

\begin{abstract}

Assembly modeling plays an important role in constructing product models, yet it is still mainly performed via interactive operations that are experience-dependent and labor-intensive. Data-driven strategies hold promise for automating this process; however, existing approaches either predict poses that cannot directly generate editable B-Rep assembly constraints or require prescribed part pairs or predefined assembly faces. %

To reduce the large candidate space of constraint prediction without predefined assembly faces, we propose a data-driven framework based on assembly feature mate prediction. The framework performs automatic assembly modeling from unassembled B-Rep parts, using assembly features to group the geometric entities of local mating interfaces and restrict candidate formation. Given the lack of suitable assembly feature annotations in existing datasets, we construct a large-scale dataset through an automatic annotation pipeline. Using the annotated dataset, we train an attention-based network to predict assembly feature mates, with Engineering Compatibility Fusion (ECF) incorporating explicit engineering compatibility into learned matching scores. To improve the accuracy of automatic assembly modeling, we use assembly feature patterns (AFPs) to guide the matching process. Finally, a post-processing search selects assembly feature mate and AFP mate candidates, whose associated constraints are generated and solved to complete assembly modeling.

Experimental results show that the framework achieves assembly constraint accuracies of 74.79\% on the six-assembly comparison set and 64.09\% on the extended test set of 16 assemblies, which includes the comparison set. With ground-truth face labels supplied as additional input, the corresponding accuracies increase to 81.78\% and 80.93\%, respectively.

\end{abstract}

\begin{keyword}
assembly feature\sep
assembly feature pattern\sep
automatic assembly modeling\sep
annotated assembly dataset\sep
attention-based networks

\end{keyword}

\end{frontmatter}

\section{Introduction}
\label{sec:introduction}
Modern products, such as aircraft and automobiles, are typically assembled from a large number of parts represented in B-Rep format. In mainstream CAD systems, a bottom-up modeling paradigm is commonly adopted. Designers first build each part independently. They then interactively select geometric entities on the parts, such as faces, edges, and vertices, and impose a series of assembly constraints between them, thereby combining the parts into a complete assembly model of the product. Although this assembly modeling process offers a high degree of freedom and flexibility, ensuring that the assembled product model faithfully reflects design intent, including the kinematic semantics of mechanisms, is often a complex and time-consuming task. Such fidelity enables the assembled product model to effectively support downstream applications, including motion simulation, interference analysis, and assembly process planning. As product structural complexity and the number of parts grow, the repeated interactions needed to specify assembly relations contribute substantially to the assembly modeling workload and limit design efficiency \cite{jones_automate_2021}. To reduce this workload, commercial CAD systems offer semi-automatic solutions, such as SolidWorks SmartMates and NX Auto Align \cite{solidworks_smartmates_overview,siemens_nx_magnetic_snap}. Although these solutions reduce repetitive manual work, they still require users to specify assembly semantics interactively.

To reduce this dependence on user interaction, data-driven automatic assembly modeling has emerged as a promising research direction. Learning-based methods most directly related to automatic assembly modeling can be broadly divided into two groups. The first predicts assembly configurations from point clouds or meshes \cite{luJigsawLearningAssemble2023, zhaoAssemblerScalable3D2025}. Although these approaches demonstrate that neural networks can capture rich geometric priors and predict plausible part poses, their outputs are not assembly constraints anchored to B-Rep entities. Consequently, they do not directly yield editable, constraint-based CAD assembly models.

The second group operates directly on B-Rep part models to infer assembly relations anchored to B-Rep entities. AutoMate \cite{jones_automate_2021} assumes a user-specified part pair and a selected face on each part, then ranks candidate mating coordinate-frame pairs and predicts mate types. JoinABLe \cite{willis_joinable_2022} removes the local region-selection requirement but still assumes a given pair of parts that form a rigid joint; it predicts the B-Rep entities defining the joint axes and uses a neurally guided search to determine the joint pose. Vergez et al. \cite{vergez_multi-part_2025} extend prediction to multi-part assemblies through state-by-state constraint generation coupled with a geometric constraint solver. Their method, however, requires predefined assembly faces. %

\begin{figure}
    \centering
    \includegraphics[width=1\linewidth]{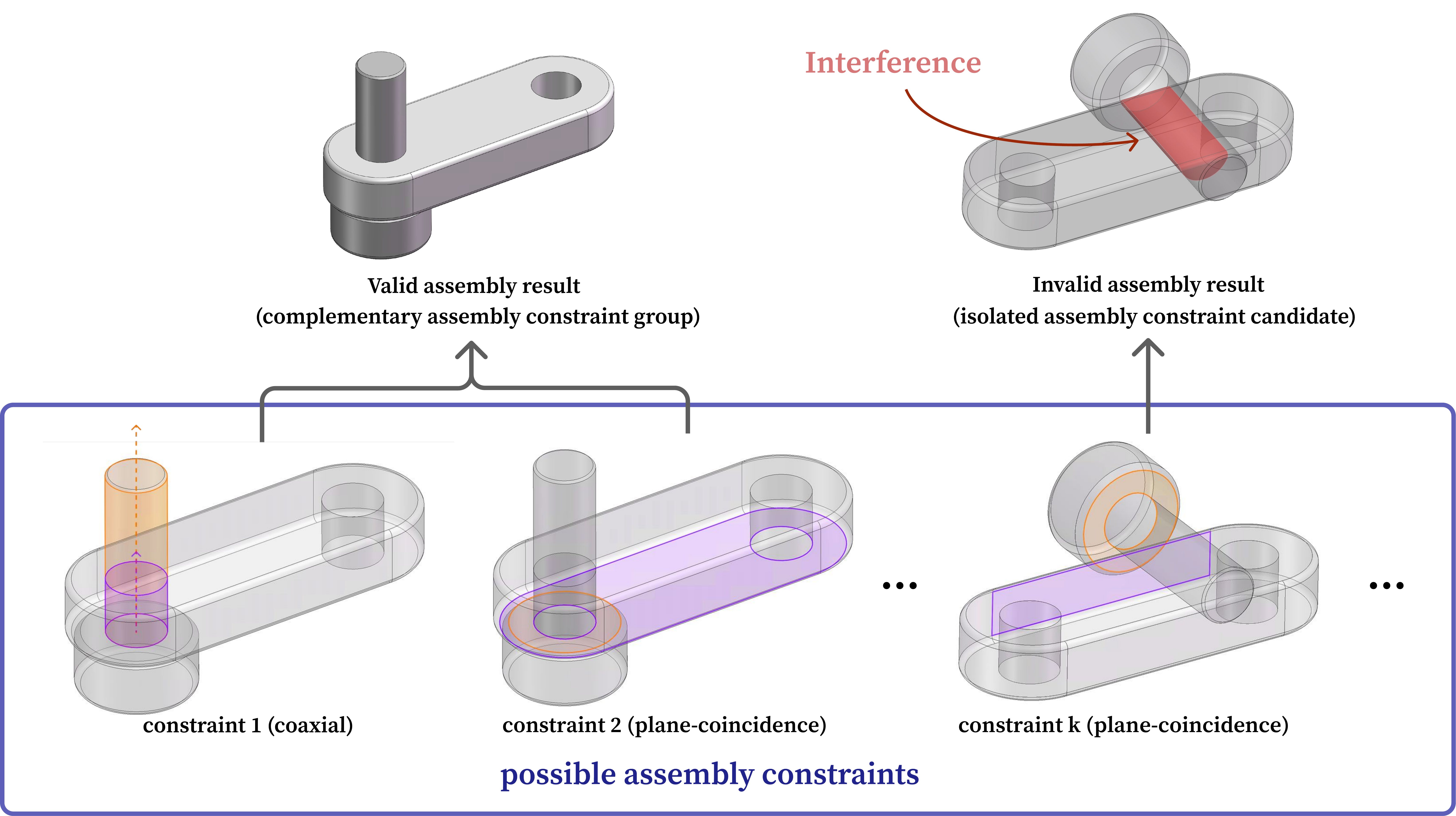}
    \caption{
Illustration of a complementary assembly constraint group and an isolated assembly constraint candidate. The orange and purple highlights distinguish the pair of geometric entities involved in each specific geometric constraint. Constraint 1, a coaxial constraint, and constraint 2, a plane-coincidence constraint, collectively define the feasible assembly result shown at the upper left. In contrast, constraint k is an isolated plane-coincidence constraint and cannot by itself constitute a candidate for the intended assembly relation, which requires a group of complementary constraints. The upper-right view shows the corresponding incorrect result, with red highlighting the geometric interference between the parts. Such complementary grouping, which characterizes the vast majority of assembly constraints, motivates our assembly feature concept.}
    \label{fig:constraint-candidate}
\end{figure}

These data-driven methods predict assembly constraints or the geometric information needed to define them. Individual assembly constraints, however, are not ideal prediction targets for automatic assembly modeling. Assembly constraints generally work in complementary groups to establish the relative position between parts, as illustrated in Fig. \ref{fig:constraint-candidate}. Without predefined assembly faces, recovering these groups requires identifying suitable combinations from a large number of geometric constraint candidates, resulting in an enormous candidate space. To make prediction feasible, existing methods reduce this space by focusing on a given part pair or using predefined assembly faces.

Overall, to tackle the prediction feasibility problem, we choose the assembly feature mate as the prediction target to build our overall framework, which is illustrated in Fig.~\ref{fig:framework}. To further improve the accuracy of automatic assembly modeling, we adopt assembly feature patterns (AFPs). To support this framework, a large-scale, high-quality dataset with rich high-level semantic annotations is essential. However, existing publicly available datasets do not contain the annotations required by our assembly feature and feature-mate formulation, although they provide valuable assembly models. Therefore, an automatic annotation pipeline is needed to construct the required dataset.

This paper makes the following contributions:
\begin{itemize}
    \item We propose a data-driven framework for automatic assembly modeling directly from unassembled B-Rep part models without predefined assembly faces.
    \item We develop an attention-based network to predict assembly feature mates, making automatic assembly modeling feasible, and incorporate AFPs to further improve the accuracy of automatic assembly modeling. %
    \item We construct a new assembly dataset through an automatic annotation pipeline. The dataset contains 529 assemblies, 43,646 parts, and 106,493 assembly features.
\end{itemize}

\section{Related work}
\subsection{Representation of Assembly Semantics in CAD}
In mainstream CAD systems, an assembly is represented by assembly constraints between B-Rep entities selected on specific part instances. Each constraint records a local relation, such as coincidence, concentricity, or distance, and a set of such constraints collectively determines the relative configuration and permitted motion of the parts. However, higher-level assembly semantics are often not explicitly represented. For example, the kinematic pair formed by two mated parts characterizes their assembly relationship in terms of the permitted relative motion and associated degrees of freedom \cite{chenMultilevelAssemblyModel2012,lupinettiAutomaticExtractionAssembly2016}, but in a constraint-based assembly model, this kinematic semantics may remain implicit in the combined effect of the related constraints. This implicit representation makes the assembly information difficult to reuse or reason about.

Higher-level representations were therefore introduced to make the assembly semantics explicit by organizing the related low-level constraints. Although their terminology and definitions are not unified \cite{rea_minango_beyond_2023}, for the purpose of CAD assembly modeling, they can be broadly organized according to semantic scope into part-side representations, which describe potential connection regions on individual parts, and cross-part representations, which associate mated interfaces with the kinematic relation and mating constraints between parts \cite{shah1993assembly,holland2000assembly,PortCompatibilityConnectabilityBased,xu_product_2013,ZHANG2021103471}. These two levels are termed assembly features and assembly feature mates, respectively, in this paper.

Feature-based approaches build on these representations by embedding part-side semantics in part models in advance, matching compatible interfaces, and instantiating the associated cross-part relations and constraints \cite{xu_product_2013,xuIntelligentAssemblyModeling2023}. Their applicability, however, depends on their predefined assembly semantics, leaving semantic acquisition as a central obstacle to broader automation.

\subsection{Recovery of Assembly Semantics from Existing Assembly Configurations}
In practical design workflows, assembly semantics---including conventional assembly constraints and higher-level assembly features---may be lost when assembly models are converted to neutral exchange formats or simplified for downstream use \cite{lupinettiAutomaticExtractionAssembly2016}; they may also be intentionally withheld when assemblies are shared to protect proprietary design knowledge. Consequently, many available assemblies preserve the part geometry and assembled configuration but lack the explicit relations needed for interpretation and reuse, motivating the recovery of assembly semantics from the remaining geometric evidence.

Starting from positioned assemblies, Lupinetti et al. analyzed interference and contact to recover degrees of freedom and candidate kinematic relations \cite{lupinettiAutomaticExtractionAssembly2016}, while related assembly/disassembly analysis derives contact and mobility information from the geometry and current configuration \cite{iacobAssemblyDisassemblyAnalysis2012}. Vergez et al. further encode planar and cylindrical contacts detected in historical ``dead'' assemblies as interface descriptors for part retrieval, recombination, and reassembly \cite{vergez_interface-based_2024}. Despite these different downstream uses, all these methods rely on prior assembly configurations in which the observed part poses already embody the assembly result. They therefore recover or reuse semantics from historical assemblies rather than determining how previously unassembled, unordered parts should be connected, and cannot provide interface information for parts without relevant assembly history.

\subsection{Learning-Based Assembly from Meshes and Point Clouds}
Learning from meshes and point clouds addresses several related geometric assembly tasks. For fragmented objects, Jigsaw predicts globally consistent poses for multiple fragments, whereas Neural Shape Mating predicts pairwise mating poses in $SE(3)$ from geometric cues \cite{luJigsawLearningAssemble2023, chenNeuralShapeMating2022}. Jigsaw++ uses learned shape priors to correct and complete a partial assembly produced by an existing assembly method \cite{luJigsawImaginingComplete2024}. For intact semantic parts, generative part-assembly methods predict part poses to reconstruct complete objects \cite{huangGenerative3DPart2020, zhang3DPartAssembly2022}, and Assembler additionally uses a reference image to guide the configuration of input part meshes \cite{zhaoAssemblerScalable3D2025}. Adjacent work on articulated-object understanding recovers complementary kinematic information: Shape2Motion infers moving parts and motion parameters from a static point cloud, while Ditto reconstructs part geometry and an articulation model from observations before and after interaction \cite{wangShape2MotionJointAnalysis2019a, jiangDittoBuildingDigital2022}.

These methods demonstrate that mesh and point-cloud geometry can support configuration and kinematic reasoning. Their prediction targets, however, are poses or articulation models rather than assembly constraints anchored to B-Rep entities. Consequently, they do not directly yield editable, constraint-based CAD assembly models.

\subsection{Learning-Based Prediction of Assembly Semantics from B-Rep Part Models}
Unlike reconstruction from existing assembly configurations, automatic assembly modeling starts from unassembled B-Rep part models without requiring predefined assembly faces on the input parts. In the absence of observed poses and contacts, the participating part pairs, mating interfaces, and assembly constraints must be inferred solely from the geometry and topology of the input parts.

Architectures such as UV-Net and BRepNet provide methods for learning representations of B-Rep parametric surface geometry and boundary topology \cite{jayaramanUVNetLearningBoundary2021, lambourneBRepNetTopologicalMessage2021}, and related B-Rep representation learning has supported shape recognition \cite{andrewr.colliganHierarchicalCADNetLearning2022, leeBRepGATGraphNeural2023} and manufacturing-feature recognition \cite{zhangBrepMFREnhancingMachining2024, wuAAGNetGraphNeural2024, ZOU2025103940}. For individual parts, Wang et al. used B-Rep graphs to identify assembly-interface faces and their kinematic types \cite{wangIntelligentIdentificationApproach2023}. This infers part-side interface semantics, but it does not determine the mating counterpart or generate cross-part assembly constraints.

For cross-part semantics, AutoMate formulates assembly modeling as an interactive recommendation task \cite{jones_automate_2021}. Given a user-specified part pair and a selected face on each part, it ranks candidate mating coordinate-frame pairs and predicts mate types. JoinABLe removes the local region-selection requirement but still assumes a specified pair of parts that form a rigid joint; it predicts the B-Rep entities defining the joint axes and uses a neurally guided search to determine the rigid joint pose \cite{willis_joinable_2022}. Both methods formulate assembly prediction for a prescribed part pair rather than inferring assembly relations over an input set of parts.

Vergez et al. address multi-part constraint prediction through state-by-state graph reasoning \cite{vergez_multi-part_2025}. At each state, their method predicts the next constraint between candidate slots and uses a geometric constraint solver to update and check the partial assembly, allowing the current assembly state and constraint cycles to be considered. The slots and their capacities, however, are extracted from contacts in prior assemblies. Thus, the input parts need not have been assembled together previously, but each part must have appeared in an assembly from which this part-side information can be obtained. The method therefore supports the recombination and reassembly of previously assembled parts for CAD database reuse, but it does not infer candidate interfaces for parts without assembly history.

Existing methods thus infer part-side semantics, complete relations for prescribed part pairs, or predict relations over a part set using predefined assembly faces. To the best of our knowledge, no existing learning-based method performs automatic assembly modeling using only the B-Rep of the input parts, without requiring predefined assembly faces at inference time.

\section{Basic concepts and approach overview}
\subsection{Basic concepts}
In this paper, assembly semantics describes mating interfaces on individual parts and assembly relations between parts. Conventional CAD systems express assembly relations through assembly constraints between B-Rep entities. To represent interfaces and their relations at higher levels, we introduce assembly features, assembly feature patterns (AFPs), and their corresponding mates below.

\subsubsection{Assembly feature and assembly feature mate} \label{sec:assembly_feature}
The concept of \textbf{assembly feature }adopted in this paper is consistent with previous works \cite{xu_product_2013, xuIntelligentAssemblyModeling2023, ZHANG2021103471}. Formally, the assembly feature ${F}$ is represented as a tuple encapsulating both geometry and semantics:
\begin{equation}
{F} = (\mathrm{GE}, {A}),
\end{equation}
where:
\begin{itemize}
    \item[1.] $\mathrm{GE} = \{e_1, e_2,..., e_n\}$ is a set of geometric entities that jointly define the geometric shape of the assembly feature.
    \item[2.] ${A} = (d_1, ..., d_m, t)$ is an attribute vector encoding the semantics of the assembly feature, where $d_1, ... ,d_m$ are the dimensional parameters (e.g., radius or width of a slot), and $t$ is a categorical variable indicating the feature type (e.g., hole, shaft, slot).

\end{itemize}

Although this representation also accommodates slots, the present study focuses on hole and shaft interfaces, motivated by the feature distributions discussed in Section~\ref{sec:dataset_results}.

Based on the concept of assembly feature, an \textbf{assembly feature mate} is introduced and expressed as
\begin{equation}
    \mathrm{FM} = ({F}, \overline{F}, {C}),
\end{equation}
where ${F}, \overline{{F}}$ are two mated assembly features from different parts that jointly form a kinematic pair; ${C} = \{c_1, c_2, ..., c_k\}$, with $k \geq 2$, denotes the set of complementary assembly constraints (e.g., face coincidence or axis alignment) that jointly realize this kinematic relation in a specific assembly configuration.

\begin{figure}
    \centering
    \includegraphics[width=1\linewidth]{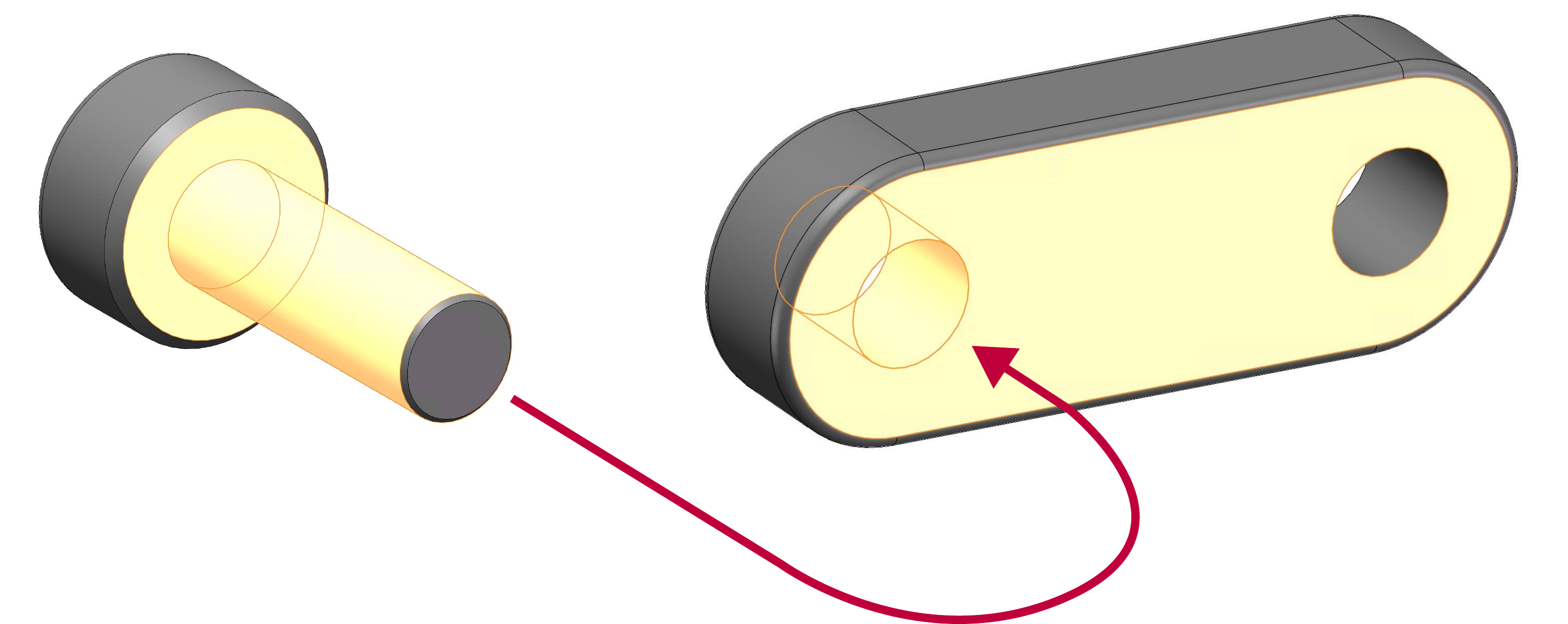}
    \caption{Illustration of assembly feature and assembly feature mate (highlighted in yellow). The assembly feature mate consists of a shaft feature (left) and a hole feature (right), each comprising a cylindrical face and a planar face. Together, they constitute an assembly feature mate, indicating that the shaft should be inserted into the hole.}
    \label{fig:AssemblyFeatureExample}
\end{figure}

Compared with the individual geometric entities referenced by assembly constraints, an assembly feature is a higher-level semantic object. In this study, we use assembly feature mates as prediction targets. Each target associates two mating features with the complementary constraints that jointly define their assembly relation, so these constraints are not treated as independent assembly decisions. This formulation also provides the explicit feature semantics used by the prediction modules described below.

\subsubsection{Assembly feature pattern and assembly feature pattern mate}\label{sec:assembly_feature_pattern}
Based on the concept of assembly feature, we then introduce the concept of assembly feature pattern. An \textbf{assembly feature pattern} (AFP), denoted as ${P}$, is a group of repetitive assembly features on a part that share unified high-level assembly semantics. It is formally defined as the tuple:
\begin{equation}
{P} = (\{ {F}_i \}_{i=1}^N,\mathrm{Pose}, \mathrm{Layout}).
\end{equation}

Here, $\{{F}_i \}_{i=1}^N$ is the set of $N$ constituent assembly features. The term $\mathrm{Pose}$ specifies the extrinsic placement of the AFP within its reference planar face and is represented by a two-dimensional origin $\mathbf{o}_P \in \mathbb{R}^2$ and a unit in-plane direction $\mathbf{r}_P \in \mathbb{R}^2$. The term $\mathrm{Layout}$ describes the intrinsic configuration of the constituent features. Each feature ${F}_i$ is assigned a representative position $\mathbf{x}_i \in \mathbb{R}^2$ in the local coordinate system of the reference plane; for a cylindrical hole or shaft, this position is obtained from the intersection of its cylinder axis with the plane. The layout is then formulated as a pairwise distance matrix \(D \in \mathbb{R}^{N \times N}\), where $D_{ij}=\lVert \mathbf{x}_i-\mathbf{x}_j\rVert_2$. This pairwise metric provides a symmetry-invariant representation of the AFP's intrinsic shape, allowing AFPs to be compared independently of their absolute in-plane positions and orientations. Its matrix representation nevertheless depends on the feature indexing: re-indexing the features by a permutation matrix $\mathbf{Q}$ transforms it to $\mathbf{Q}D\mathbf{Q}^{T}$ without changing the underlying pairwise metric structure.

\begin{figure}[htbp]
  \centering
  \includegraphics[width=0.8\linewidth]{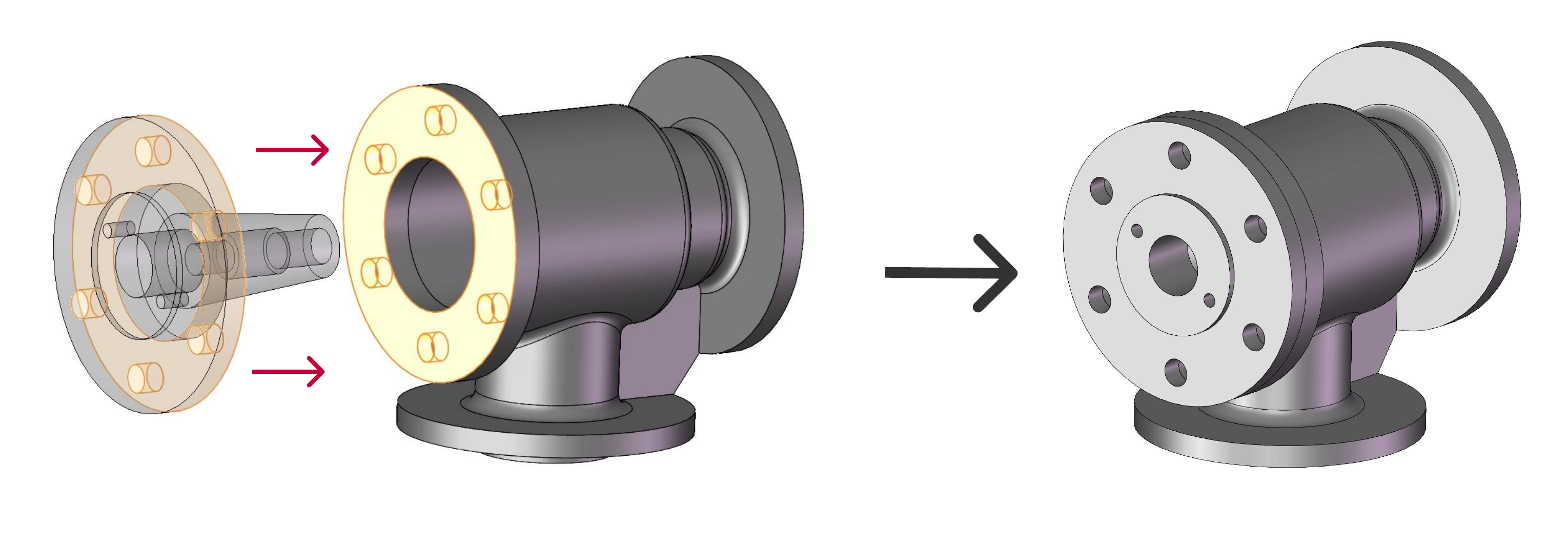}
  \caption{Illustration of assembly feature patterns and an AFP mate. Left: the highlighted AFPs on the two parts each consist of six hole features arranged in a circle. The AFP mate relates the two patterns through a layout-consistent feature correspondence. Right: the assembled configuration.}
  \label{fig:AssemblyFeaturePattern}
\end{figure}

Figure \ref{fig:AssemblyFeaturePattern} illustrates two AFPs, each consisting of six hole features arranged in a circle on one part. To establish the illustrated assembly relation, the constituent features must be matched jointly according to the layouts of the two AFPs. Building on this idea, an \textbf{assembly feature pattern mate} (AFP mate), denoted as $\mathrm{PM}$, is defined as a layout-consistent relation between two AFPs. It is formally represented as
\begin{equation}
    \mathrm{PM} = (P, \overline{P}, M, \Gamma),
\end{equation}
where $P$ and $\overline{P}$ contain $N$ and $\overline{N}$ assembly features, respectively. Without loss of generality, let $\overline{N}\leq N$. The correspondence $M$ is induced by an injective mapping from the smaller AFP to the larger AFP:
\begin{equation}
    \pi: \{1,\ldots,\overline{N}\} \hookrightarrow \{1,\ldots,N\},
    \qquad
    M = \{(\pi(j),j) \mid j=1,\ldots,\overline{N}\}.
\end{equation}
Thus, every feature of the smaller AFP participates in the correspondence, whereas unmatched features may remain in the larger AFP. The term $\Gamma$ records the assembly constraint set associated with each feature match in $M$. Specifically, for every $(i,j)\in M$, $\Gamma(i,j)$ denotes the set of assembly constraints that collectively define the assembly feature mate between $F_i$ in $P$ and $\overline{F}_j$ in $\overline{P}$.

The AFP represents a higher-level semantic object compared with a single assembly feature, and the AFP mate extends this abstraction to relations between feature groups. Its correspondence $M$ is established from the structural layouts of $P$ and $\overline{P}$ and evaluates the constituent feature pairs jointly. This group-level relation can therefore exclude locally plausible feature correspondences that are inconsistent with the overall layout.

\subsection{Overall approach}
To automate the assembly modeling process from B-Rep part models without predefined assembly faces, we present a three-module framework centered on assembly feature mate prediction and AFP mate candidate generation. The overall framework combines: (1) an automatic semantic extraction pipeline for constructing a semantically rich dataset, (2) an attention-based network that uses parent-part context for assembly feature mate prediction, and (3) a deterministic mechanism for AFP recognition and groupwise correspondence generation. Before introducing each component in detail, three key issues and their corresponding solutions are described below:
\begin{itemize}
    \item [(1)] \textbf{How to automatically construct a high-quality dataset with rich semantic annotations?}
    To facilitate the construction of a new large-scale dataset with rich annotations of assembly features and assembly feature patterns, we propose an automatic semantic extraction pipeline. This pipeline identifies and reconstructs the geometric constraints in the CAD assembly models, subsequently transforming them into high-level assembly features and pattern annotations. The resulting dataset provides the essential supervision required by the semantics-aware learning in our framework.

    \item [(2)] \textbf{How to represent and evaluate assembly relations without treating constraints independently?}
    We shift the prediction target from individual low-level geometric constraints to assembly feature mates, which associate mating features with groups of complementary constraints. The assembly feature mate prediction network uses Engineering Compatibility Fusion to combine learned matching evidence with explicit engineering compatibility. At the next semantic level, AFP recognition and groupwise correspondence generation produce layout-consistent AFP mate candidates that can exclude locally plausible but globally inconsistent feature correspondences.

    \item [(3)] \textbf{How to organize inference over a large candidate space?}
Each assembly feature mate represents a group of complementary constraints, so the framework avoids treating every constituent constraint as a separate assembly decision. Part embeddings provide context for scoring assembly feature mate candidates. The resulting assembly feature mate and AFP mate candidates are ranked by their scores.
\end{itemize}

\begin{figure}[htbp]
    \centering
    \includegraphics[width=1\textwidth]{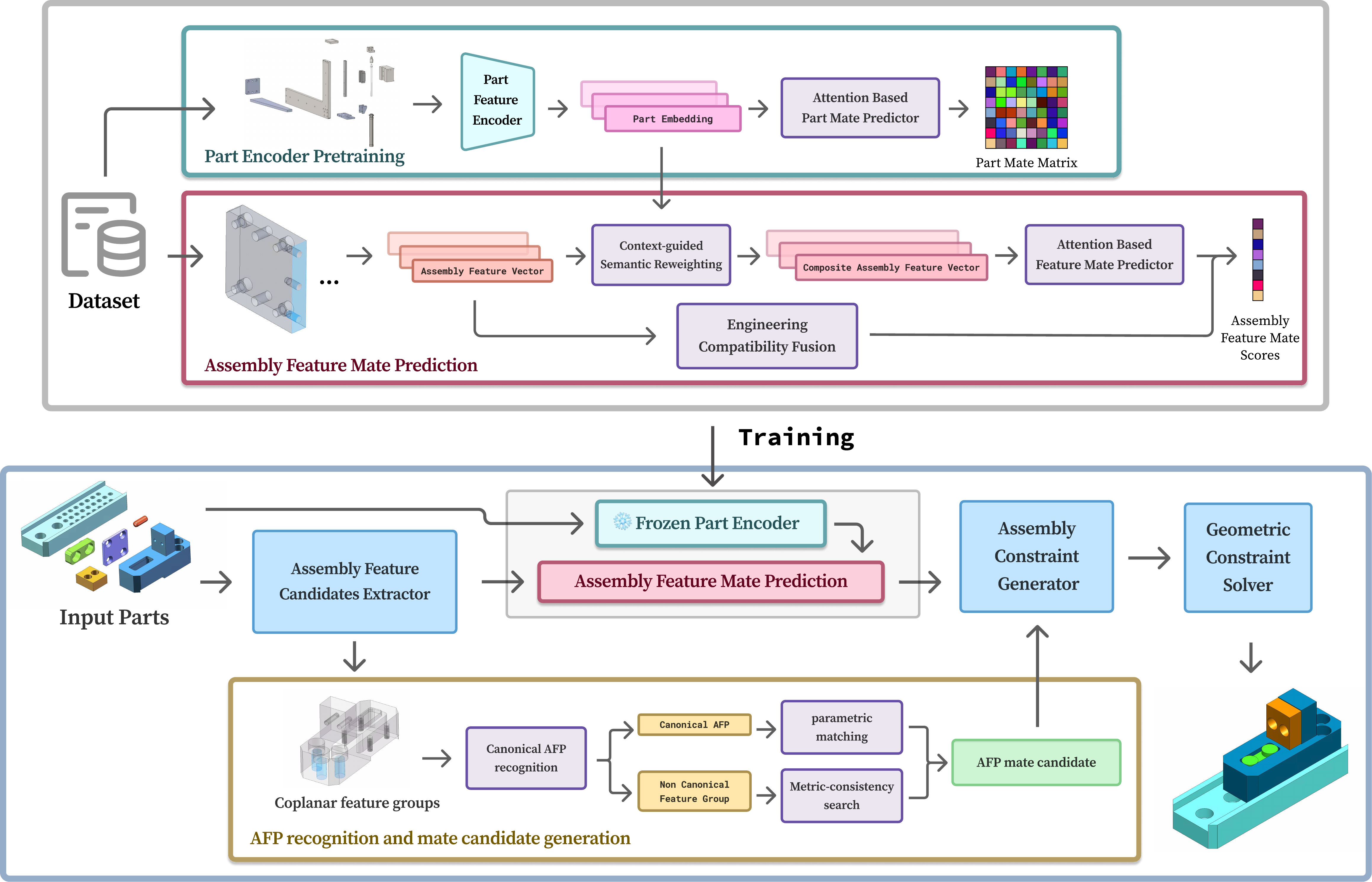}
    \caption{Overall architecture of the proposed approach. The framework consists of a training stage (top gray box) and an assembly modeling stage (bottom flow). The part encoder is first pretrained through part mate prediction and then frozen during assembly feature mate prediction training. During assembly modeling, the encoder provides part embeddings as context for assembly feature mate prediction, which also uses the extracted assembly feature candidates. Coplanar feature groups enter the AFP recognition and mate candidate generation module, which uses parametric matching for canonical AFP hypotheses and metric-consistency search for groups that cannot be assigned a canonical layout type. Both paths output AFP mate candidates. The assembly constraint generator uses the scored assembly feature mate candidates and AFP mate candidates to select assembly relations and instantiate assembly constraints. The geometric constraint solver processes these constraints to obtain the assembly configuration.}
    \label{fig:framework}
\end{figure}

Based on the above solutions, we designed a framework of our approach, which is illustrated in Fig. \ref{fig:framework}. It consists of a training stage that utilizes labeled data and an assembly modeling stage that operates on B-Rep CAD models without predefined assembly faces.

\textit{Training stage.} Training proceeds in two stages using annotations from the constructed dataset:
\begin{itemize}
    \item During \textbf{part encoder pretraining}, a UV-Net-based encoder extracts part embeddings, which are fed into an attention-based part mate predictor to construct a part mate matrix. The part mate prediction task provides supervision for pretraining the encoder.
    \item For \textbf{assembly feature mate prediction}, the pre-trained part encoder is frozen. The module combines part and feature embeddings to form composite assembly-feature embeddings and estimates learned matching scores for candidate features. An Engineering Compatibility Fusion (ECF) mechanism then combines these scores with explicit engineering compatibility to produce the final assembly feature mate scores.
\end{itemize}

\textit{Assembly modeling stage.} In this stage, the input consists of raw B-Rep CAD models without predefined assembly faces.
\begin{itemize}
    \item The stage begins with assembly feature candidate extraction based on topological adjacency traversal. For hole and shaft features, the algorithm iterates over cylindrical faces and locates associated planar faces. Each cylindrical face is combined with an associated planar face to form an assembly feature candidate, excluding transitional faces such as chamfers and fillets from its composition. Extraction prioritizes avoiding omissions, so the retained candidates may include geometric combinations that are not assembly features. The extracted candidates are then grouped by their reference planes into coplanar feature groups that provide the search regions for AFP recognition.
\end{itemize}
 \begin{itemize}
    \item The frozen part encoder maps the input parts to part embeddings. The \textbf{assembly feature mate prediction} module uses these embeddings as context together with the extracted assembly feature candidates to predict assembly feature mate scores. The \textbf{AFP recognition and mate candidate generation} module receives the coplanar feature groups, recognizes rectangular and circular AFP hypotheses, and generates their correspondences from layout parameters. Metric-consistency search handles feature groups that cannot be assigned either layout type. Both paths produce AFP mate candidates for the assembly constraint generator.
    \item Finally, an \textbf{assembly constraint generator} synthesizes the outputs from the preceding modules by ranking both assembly feature mate and AFP mate candidates based on their calculated confidence scores. The associated assembly constraints are applied during part installation and processed by the geometric constraint solver. During installation, each candidate is checked for conflicts with the existing assembly constraints and for geometric interference between the two parts involved; only candidates that pass these checks are retained.

The details of the above-mentioned modules are described in the following sections.
\end{itemize}

\section{Dataset construction}
\label{sec:dataset}
To support semantics-aware data-driven automatic assembly modeling, we adopt high-level semantic targets, including assembly features, assembly feature mates, assembly feature patterns, and AFP mates. However, existing open datasets primarily focus on assembly constraint or part position annotations. Manually annotating these high-level semantic targets is labor-intensive and impractical for research groups. Hence, we propose an automatic annotation method to build assembly feature and assembly feature pattern annotations from assemblies with assembly constraint data. To achieve this, two key issues need to be solved.

\begin{figure}[htbp]
  \centering
  \includegraphics[width=0.8\linewidth]{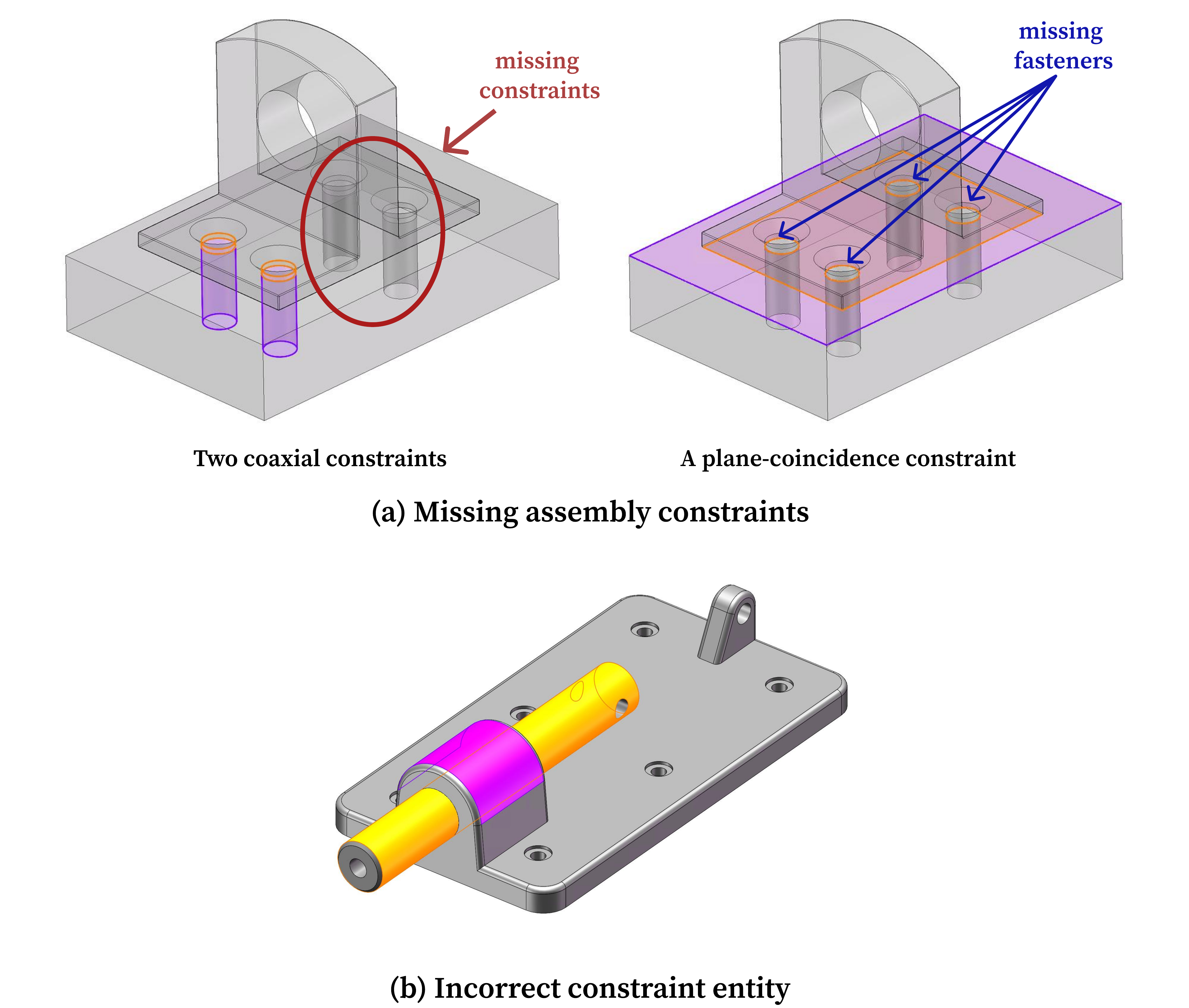}
  \caption{Illustration of missing and incorrect assembly constraints in assembly models. The orange and purple highlights indicate the geometric entities referenced by the assembly constraints. (a) Missing assembly constraints: The two parts are assembled using two coaxial assembly constraints and a plane-coincidence assembly constraint. The fasteners of four holes are omitted, leaving coaxially aligned holes in the simplified assembly. Moreover, since the three given assembly constraints have already fully restricted the degrees of freedom, no assembly constraint is annotated on the other two holes (red circle). (b) Incorrect assembly constraint entity: To align the pin, the user selected the external cylindrical surface (highlighted in purple) instead of the mechanically correct inner bore. While this selection shares the same axis and satisfies the geometric constraint solver, it misrepresents the physical semantics of the assembly.}
  \label{fig:Ambiguous and Under-specified}
\end{figure}

The first issue is the absence of some constraints. This issue has two main sources.
The first source is common simplification practices in assembly modeling, where standard fasteners are frequently omitted (as shown in Fig. \ref{fig:Ambiguous and Under-specified}(a)). This absence of parts leads to implicit 'hole-to-hole' alignments rather than well-defined geometric constraints. The second source is the simplification of the user-defined constraints themselves. This is because users typically define only the minimum constraints required for spatial positioning, frequently omitting the full set of geometric constraints that describe the actual physical connection. The physical reality of an assembly, however, always contains numerous redundant constraints, as shown in Fig. \ref{fig:Ambiguous and Under-specified}(a).

The second issue stems from the geometric entities in user-defined geometric constraints. In CAD systems, an assembly constraint can be equivalently defined using various topological elements, such as edges, vertices, or faces. Moreover, besides the type of geometric entities, the selection is still not unique. As illustrated in Fig. \ref{fig:Ambiguous and Under-specified}(b), a user may select the cylindrical face that is largest or easiest to click in the graphical user interface (GUI), even if it is not the actual surface that makes physical contact in the final assembly. The resulting assembly constraint is geometrically valid but semantically misleading, as it does not reflect the real interaction.

To address the issues, we developed an automatic dataset annotation pipeline.

\subsection{Assembly feature and assembly feature mate types}
\label{sec:canon_mates}
This study focuses on hole and shaft features, which account for the vast majority of assembly features in the constructed dataset, as reported in the Results section. The corresponding \textbf{hole-shaft} assembly feature mates describe direct connections between these feature types. Furthermore, empirical analysis of real-world CAD assemblies reveals that designers frequently omit standard fasteners, such as bolts and pins, to enhance computational performance and simplify the assembly models, as illustrated in Fig. \ref{fig:Ambiguous and Under-specified}(a). Given this industrial practice, we formalize the \textbf{hole-hole} correspondence as a legitimate assembly feature mate type. The present study therefore emphasizes hole-shaft and hole-hole assembly feature mates.

\subsection{Automatic dataset annotation pipeline}
\label{sec:canonicalization_pipline}
\begin{figure}[htbp]
  \centering
  \includegraphics[width=1\linewidth]{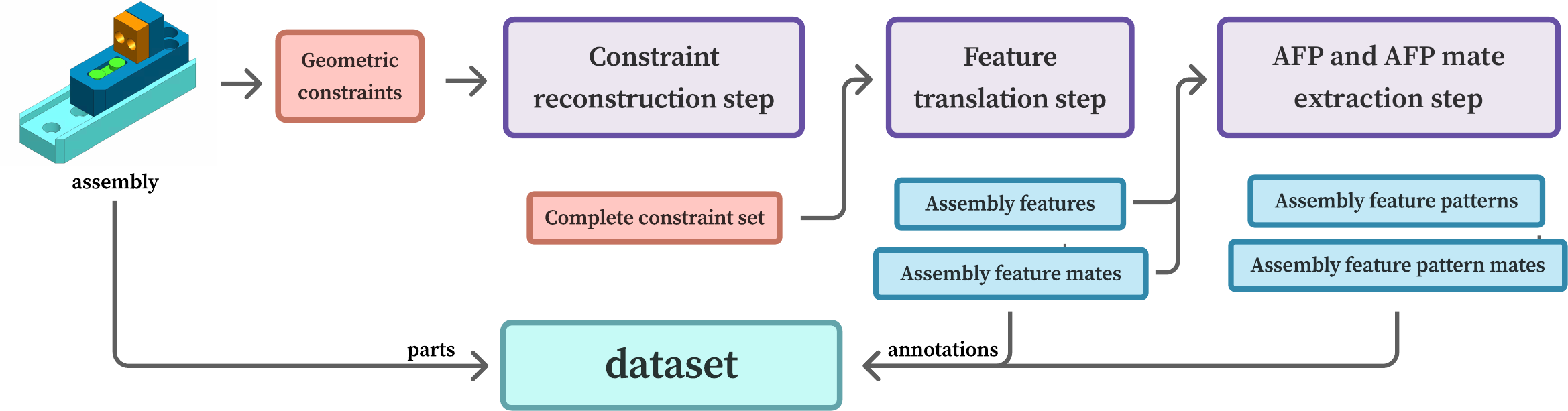}
  \caption{The automatic annotation pipeline.}
  \label{fig:dataset}
\end{figure}

To construct the required dataset, our pipeline transforms the low-level, ambiguous geometric constraints into standardized and semantically rich targets through the following three steps, which are illustrated in Fig. \ref{fig:dataset}.

The pipeline begins with a constraint reconstruction step to complete the omitted geometric constraints as well as correct them. During this step, the pipeline evaluates the validity of each constraint by checking for entity coincidence and the presence of associated partners. If a constraint is found to be incomplete or misaligned, the pipeline leverages geometric adjacency-based search to retrieve the correct geometric entities. This corrective step serves as a critical prerequisite for the subsequent identification of assembly features and assembly feature mates.

Next, the feature translation step is applied to convert all user-defined constraints into a uniform, face-based assembly feature target. Our pipeline resolves the discrepancy of geometric constraints by systematically translating these non-face constraints into their underlying face-based equivalents. This stage also serves as a critical validation step: by checking for physical contact between the resolved faces, the system can identify and filter out semantically misleading constraints. The output is a set of assembly features and assembly feature mates.

The final assembly feature pattern and AFP mate extraction stage involves the recognition of assembly feature patterns and their corresponding AFP mates. The pipeline begins by grouping all kinematically complete assembly feature mates that share a common spatial context, such as all hole-shaft assembly feature mates whose features reside on the same planar surface of a part. Then, the mate relationships of the grouped assembly features are extracted to construct the AFP mates. Though these pattern-level annotations are not utilized to train the neural network in the current study, they are preserved to lay the groundwork for future research on AFP mate prediction.

This process standardizes existing annotations and recovers omitted mating relations for the interface types considered in this work, providing annotated data for subsequent network training.

\section{Part encoder pretraining}
\label{sec:part_relation}
\begin{figure}[htbp]
    \centering
    \includegraphics[width=1\linewidth]{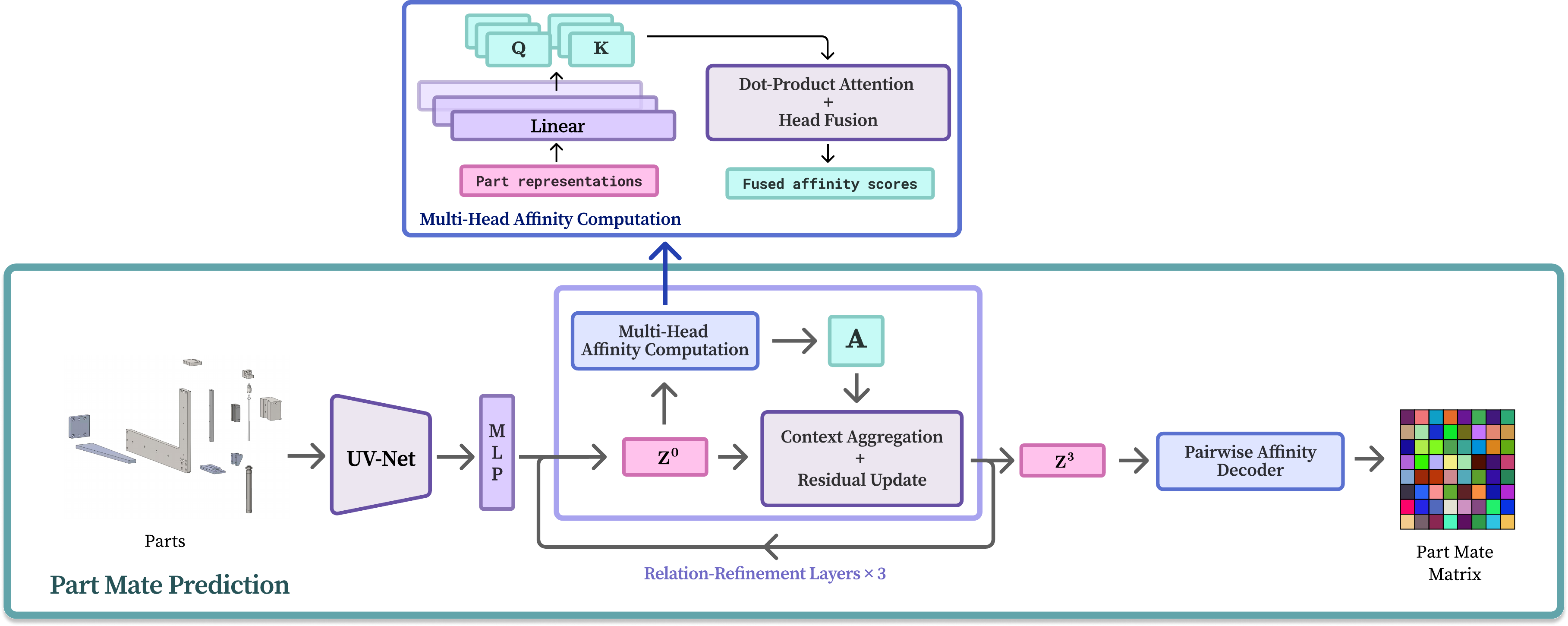}
    \caption{Part encoder pretraining through part mate prediction.}
    \label{fig:part_level}
\end{figure}

We pretrain a part encoder through part mate prediction, an assembly-related task, to provide context for subsequent assembly feature mate prediction. Using individual B-Rep CAD models as input, the encoder learns part embeddings that capture the geometric and topological properties of the parent parts. The module is illustrated in Fig. \ref{fig:part_level}.

\subsection{Part embedding extractor}
Each part is first encoded using a UV-Net-based encoder \cite{jayaramanUVNetLearningBoundary2021}, which captures fine-grained geometric and topological information from the B-Rep representation. This encoder integrates three modules—a curve encoder, a surface encoder, and a graph encoder—that together generate an embedding vector $\mathbf{p}_i \in \mathbb{R}^{d}$ for each part.

These part embeddings serve as the input for both the part mate prediction and assembly feature mate prediction modules. In our implementation, the embedding dimension is set to $d = 128$.

\subsection{Part mate prediction}
\label{subsec:part-level correlation}
Let $N$ be the number of parts in an assembly. Their UV-Net embeddings are stacked row-wise to form
\begin{equation}
\mathbf{X} = [\mathbf{p}_1^\top;\ldots;\mathbf{p}_N^\top] \in \mathbb{R}^{N \times d}.
\end{equation}
A shared two-layer MLP, with ReLU activation and a dropout rate of 0.2 after each layer, projects these embeddings into the initial latent representation
\begin{equation}
\mathbf{Z}^{(0)} = \mathrm{MLP}_{\mathrm{in}}(\mathbf{X}) \in \mathbb{R}^{N \times d}.
\end{equation}

The latent representation is subsequently refined by $L=3$ relation-refinement layers with identical architectures but independent parameters. For layer $l \in \{1,\ldots,L\}$, the input is $\mathbf{Z}^{(l-1)}$. Pairwise affinities are computed from $H=16$ sets of query and key projections, with $d_k=d/H=8$. For head $h$, the projected representations are
\begin{equation}
\begin{aligned}
\mathbf{Q}^{(l,h)} &= \mathbf{Z}^{(l-1)}\mathbf{W}_Q^{(l,h)}, \\
\mathbf{K}^{(l,h)} &= \mathbf{Z}^{(l-1)}\mathbf{W}_K^{(l,h)}.
\end{aligned}
\end{equation}
The affinity distribution for each head is computed by scaled dot-product attention:
\begin{equation}
\mathbf{A}^{(l,h)} =
\operatorname{softmax}_{j}\left(
\frac{\mathbf{Q}^{(l,h)}(\mathbf{K}^{(l,h)})^\top}{\sqrt{d_k}}
\right) \in \mathbb{R}^{N \times N},
\end{equation}
where the softmax is applied over the key-part index $j$ for each query part $i$. Thus, $A_{ij}^{(l,h)}$ quantifies the relative contribution of part $j$ when updating the representation of part $i$ in head $h$. The head-specific matrices are combined through a learnable linear projection:
\begin{equation}
\mathbf{A}^{(l)} = \sum_{h=1}^{H}\beta_h^{(l)}\mathbf{A}^{(l,h)},
\end{equation}
where $\beta_h^{(l)}$ denotes the learnable fusion weight for head $h$. The resulting matrix $\mathbf{A}^{(l)} \in \mathbb{R}^{N \times N}$ is an intermediate relational representation used to aggregate context among the parts, rather than a part mate probability matrix. In each refinement layer, the fused affinity matrix $\mathbf{A}^{(l)}$ directly weights the current part representations $\mathbf{Z}^{(l-1)}$, which serve as the value features without an additional projection. The layer output is obtained through context aggregation, a learnable linear update, and a residual connection:
\begin{equation}
\mathbf{Z}^{(l)}
= \mathrm{ReLU}\left(\mathbf{Z}^{(l-1)}
 + \mathrm{Linear}_\mathrm{upd}^{(l)}\left(
 \mathbf{A}^{(l)}\mathbf{Z}^{(l-1)}\right)\right).
\end{equation}

After the three refinement layers, an independently parameterized pairwise affinity decoder maps $\mathbf{Z}^{(3)}$ to a raw pairwise score matrix. For each decoder head $h$, the query and key representations and their corresponding affinity matrix are computed as
\begin{equation}
\begin{aligned}
\mathbf{Q}_{\mathrm{dec}}^{(h)}
&= \mathbf{Z}^{(3)}\mathbf{W}_{Q,\mathrm{dec}}^{(h)}, \\
\mathbf{K}_{\mathrm{dec}}^{(h)}
&= \mathbf{Z}^{(3)}\mathbf{W}_{K,\mathrm{dec}}^{(h)}, \\
\mathbf{A}_{\mathrm{dec}}^{(h)}
&= \operatorname{softmax}_{j}\left(
\frac{\mathbf{Q}_{\mathrm{dec}}^{(h)}
(\mathbf{K}_{\mathrm{dec}}^{(h)})^\top}{\sqrt{d_k}}
\right).
\end{aligned}
\end{equation}
The decoder parameters are independent of those in the refinement layers. The head-specific affinity matrices are linearly combined to obtain the raw pairwise score matrix
\begin{equation}
\mathbf{R}
= \sum_{h=1}^{H}\beta_{\mathrm{dec},h}\mathbf{A}_{\mathrm{dec}}^{(h)},
\end{equation}
where $\beta_{\mathrm{dec},h}$ denotes the learnable fusion weight of decoder head $h$. Unlike the refinement layers, the decoder directly returns $\mathbf{R}$ without performing feature aggregation. A sigmoid activation converts these scores into an intermediate probability matrix, which is subsequently symmetrized because the part mate relation is undirected:
\begin{equation}
\begin{aligned}
\widetilde{\mathbf{C}} &= \operatorname{sigmoid}(\mathbf{R}), \\
\mathbf{C} &= \frac{1}{2}\left(\widetilde{\mathbf{C}}
    + \widetilde{\mathbf{C}}^\top\right).
\end{aligned}
\end{equation}
Thus, each $c_{ij} \in [0,1]$ represents the predicted probability that parts $i$ and $j$ are mated, and the final prediction strictly satisfies $c_{ij}=c_{ji}$.

\subsection{Loss function}
The part mate prediction module is trained as a binary score predictor, aligning with our "coarse-grained" affinity task. For each pair of parts $(i, j)$ in an assembly, the decoder output is converted into the symmetrized probability $c_{ij} \in [0, 1]$ defined above. The model is trained by computing the standard Binary Cross-Entropy (BCE) Loss between the predicted probabilities $\mathbf{C}$ and the ground-truth matrix $\hat{\mathbf{C}}$, where $c_{ij}$ and $\hat{c}_{ij}$ denote the $(i,j)$-th elements of $\mathbf{C}$ and $\hat{\mathbf{C}}$ respectively:
\begin{equation}
    \mathcal{L}_\mathrm{part} = \frac{1}{N^2} \sum_{i,j} \text{BCE}(c_{ij}, \hat{c}_{ij})
\end{equation}
\begin{equation}
    \text{BCE}(p, \hat{p}) = - [ \hat{p} \log(p) + (1-\hat{p}) \log(1-p) ]
\end{equation}

The pre-trained part embeddings provide assembly context for the following assembly feature mate prediction module, as discussed in Section \ref{sec:context-fusion}. The part encoder remains frozen during feature-level training.

\section{Assembly feature mate prediction}
Using the pre-trained part embeddings from Section \ref{sec:part_relation}, this module assigns scores to candidate pairs of assembly features. Each candidate assembly feature mate is treated as a unified relation rather than as a set of independent assembly-constraint decisions. To achieve accurate assembly feature mate prediction, three key issues must be addressed. First, the functional role of a locally defined assembly feature depends not only on its local geometry but also on the context of its parent part. Second, each query feature must be evaluated against multiple candidate features within a common prediction space. Third, valid assembly feature mates are governed by strict engineering semantics, which must be incorporated into probabilistic matching to improve prediction accuracy.

\begin{figure}[htbp]
    \centering
    \includegraphics[width=0.8\linewidth]{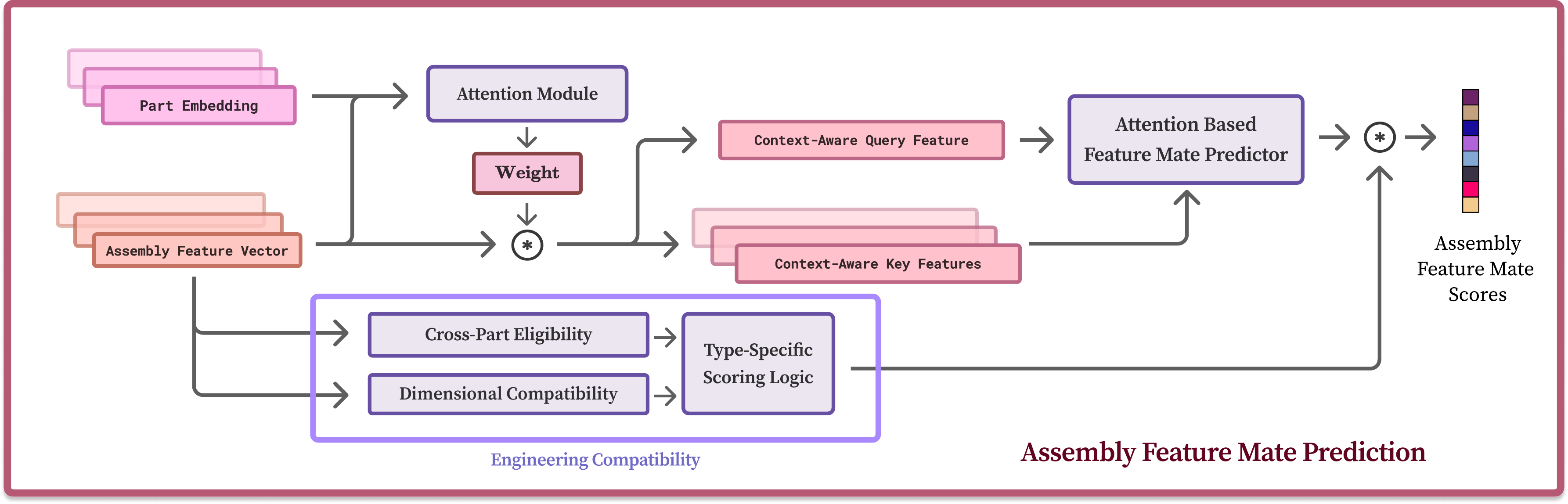}
    \caption{Framework of assembly feature mate prediction. The attention-based feature mate predictor estimates learned matching scores, and ECF combines them with cross-part eligibility and type-conditioned dimensional compatibility to obtain compatibility-adjusted assembly feature mate candidate scores.}
    \label{fig:feature_level}
\end{figure}

To address these issues, the proposed module follows a context-aware and engineering-guided prediction process. First, a context-guided semantic reweighting mechanism uses parent-part embeddings to guide the aggregation of local candidate feature representations. Second, an attention-based feature mate predictor estimates learned matching scores for the candidate set. Finally, an Engineering Compatibility Fusion (ECF) mechanism incorporates explicit engineering compatibility into these scores to produce compatibility-adjusted candidate scores. The overall framework is illustrated in Fig. \ref{fig:feature_level}.

\subsection{Assembly feature vector extraction}
In this work, each hole or shaft assembly feature $F$ is extracted into a unified 21-dimensional raw vector $\mathbf{f}$. This vector serves as the initial input to the neural network modules. Specifically, it includes the feature's semantic type (hole or shaft), key geometric parameters, and its position and orientation relative to the part's coordinate system. As detailed in Table \ref{tab:feature_vector_detailed}, each vector begins with a 2-dimensional header containing part and type identifiers. The subsequent 19 dimensions constitute the feature body, formed by concatenating 8-dimensional cylinder parameters and 6-dimensional plane parameters, followed by 5-dimensional zero-padding.

The part identifier, feature-type identifier, and primary mating dimension are also used to evaluate the explicit engineering compatibility introduced in the ECF mechanism.

\begin{table}[htbp]
    \centering
    \caption{Composition of the 21-dimensional assembly feature vector ($\mathbf{f}$) for hole and shaft features.}
    \label{tab:feature_vector_detailed}
    \scalebox{1}{
        \begin{tabular}{l l c p{7cm}}
            \toprule
            \textbf{Type} & \textbf{Component} & \textbf{Dim.} & \textbf{Content / Description} \\
            \midrule
            \multirow{2}{*}{\textbf{All}}
            & Part ID ($\text{id}_p$)& 1 & Hash of the part's identifier. \\
            & Type ID ($\text{id}_t$) & 1 & Integer for feature type. \\
            \midrule
            \textbf{Hole/Shaft} & \multicolumn{3}{l}{\textbf{Feature Body} (Total 19 dims)} \\
            &  Cylinder Params ($\mathbf{f}_\mathrm{cyl}$)& 8 & Concatenation of radius, axis, origin, and sense: $(r, \mathbf{v}_{\text{axis}}, \mathbf{p}_{\text{org}}, s)$. \\
            &  Plane Params ($\mathbf{f}_\mathrm{pl}$)& 6 & Concatenation of normal and origin: $(\mathbf{v}_{\text{norm}}, \mathbf{p}_{\text{org}})$. \\
            &  \textbf{Padding} & 5 & Zero-padding to reach the total length. \\
            \bottomrule
        \end{tabular}
        }
\end{table}

\subsection{Context-guided semantic reweighting mechanism}

\label{sec:context-fusion}
The local feature vectors extracted above are combined with parent-part embeddings from the B-Rep encoder in Section~\ref{sec:part_relation} to construct representations for assembly feature mate prediction. We use a \textbf{context-guided semantic reweighting} mechanism in which parent-part information guides the weighted aggregation of candidate feature representations.

For a candidate set of $N$ assembly features, let $\mathbf{f}_i$ denote the local vector of candidate $i$ and $\mathbf{p}_i$ its parent-part embedding. Multiple candidates may belong to the same part. Each local vector provides a query and value, while its parent-part embedding provides the corresponding key. Linear projection and layer normalization give column vectors $\mathbf{q}_i,\mathbf{k}_i,\mathbf{v}_i\in\mathbb{R}^{d_k}$, with $d_k=128$. Each local query is compared with all $N$ parent-part keys, and the resulting scores are normalized over the key index $j$ within the candidate set. The reweighted representation is the corresponding weighted sum of local values:

\begin{equation}
\begin{aligned}
\{\mathbf{q}_i, \mathbf{k}_i, \mathbf{v}_i\} &= \text{LayerNorm}(\{\mathbf{W}_Q \mathbf{f}_i, \mathbf{W}_K \mathbf{p}_i, \mathbf{W}_V \mathbf{f}_i\}), \\
\hat{\mathbf{F}} &= \operatorname{Softmax}_{j}\left(\frac{\mathbf{Q}\mathbf{K}^\top}{\sqrt{d_k}}\right)\mathbf{V}.
\end{aligned}
\end{equation}

Here, $\mathbf{Q},\mathbf{K},\mathbf{V}\in\mathbb{R}^{N\times d_k}$ stack the projected vectors row-wise, and row $i$ of $\hat{\mathbf{F}}$ is $\mathbf{\hat{f}}_i^\top$. Softmax is applied row-wise over the key index $j$ of the $N\times N$ score matrix. Each $\mathbf{\hat{f}}_i$ is a weighted average of the local candidate values, with weights determined by the interactions between its local query and the candidate parent-part keys. In this way, assembly context from the parent parts guides the aggregation of candidate feature information for subsequent assembly feature mate prediction.

\subsection{Attention-based feature mate predictor}
The predictor uses candidate feature embeddings that incorporate assembly context through the reweighting mechanism in Section~\ref{sec:context-fusion}. A multi-head attention mechanism then estimates the relative matching evidence between a query feature and its candidates. In a single forward pass, it assesses a query feature against a candidate set and allows the candidates to compete within a common prediction space. During training, each query is paired with one positive candidate and 19 negative candidates sampled from the global set of assembly features. During assembly modeling, the extracted candidate features for each query are partitioned into groups of 20 and scored separately within each group. The candidate set size is limited to $N = 20$ because encoding the source parts, which may comprise hundreds of faces, imposes substantial memory requirements. Larger candidate sets would therefore require additional memory or simpler part geometries.

For a given feature $\mathbf{\hat{f}}_{q}$, it serves as the query. Its sampled or constructed candidate set is denoted by $\mathcal{C}_q=\{c_1,\ldots,c_N\}$, and the corresponding $N$ feature vectors are stacked column-wise as $\mathbf{F}_\text{candidate} = [\mathbf{\hat{f}}_{c,1}, ..., \mathbf{\hat{f}}_{c,N}]$.

The learned matching score for each feature pair is computed through a multi-head attention-based fusion mechanism, as in Section \ref{subsec:part-level correlation}. For each attention head $h \in \{1,\ldots,H\}$, the attention is computed as follows:

\begin{align}
    \mathbf{q}^{(h)} &= \mathbf{W}_q^{(h)}\mathbf{\hat{f}}_{q} \\
    \mathbf{K}^{(h)} &= \mathbf{W}_k^{(h)} \mathbf{F}_\text{candidate} \\
    \mathbf{e}^{(h)} &= \frac{(\mathbf{K}^{(h)})^\top\mathbf{q}^{(h)}}{\sqrt{d_k}}
\end{align}

Here, $\mathbf{W}_q^{(h)}$ and $\mathbf{W}_k^{(h)}$ are the learnable weight matrices for the query and key projections of head $h$. The scores within each head are normalized over the candidate dimension by Softmax. For each candidate, the resulting head-specific weights are concatenated and passed through a shared output projection to generate the corresponding entry of $\boldsymbol{\alpha}\in\mathbb{R}^{N}$:
\begin{equation}
    \boldsymbol{\alpha} = \mathrm{MLP}([\operatorname{softmax}(\mathbf{e}^{(1)}), ... ,\operatorname{softmax}(\mathbf{e}^{(H)})]).
\end{equation}

A Sigmoid activation produces the learned matching scores
\begin{equation}
    \mathbf{p}_{\theta}=\operatorname{sigmoid}(\boldsymbol{\alpha}).
\end{equation}
Here, $p_{\theta,c}=p_\theta(c\mid q,\mathcal{C}_q)$ denotes the learned matching score assigned to candidate $c$. These scores capture evidence learned from feature geometry and part context and are not normalized to sum to one across candidates. Explicit engineering semantics are incorporated in the subsequent ECF stage.

\subsection{Engineering Compatibility Fusion}

The attention-based feature mate predictor estimates relative matching evidence from data, whereas valid assembly feature mates must also satisfy engineering semantics known independently of the training data. We therefore introduce Engineering Compatibility Fusion (ECF), which represents these semantics as an engineering compatibility potential and combines it with the learned matching scores. For a query feature ${F}_q$ and a candidate feature ${F}_c$, the potential is constructed from cross-part eligibility, dimensional compatibility, and type-specific scoring logic.

\begin{itemize}
    \item [1.] \textbf{Cross-Part Eligibility}: An assembly feature mate must connect features from different parts. This condition is particularly relevant during training because random sampling from the global set of assembly features can place features from the query feature's parent part in the same candidate set. Let $h_q$ and $h_c$ denote the part identifiers associated with the query and candidate features. Cross-part eligibility is represented by the binary indicator
    \begin{equation}
        m_\mathrm{part}(q,c)=\mathbb{I}[h_q\neq h_c].
    \end{equation}
    Thus, same-part candidates receive zero compatibility, whereas cross-part candidates remain eligible for subsequent scoring.

    \item [2.] \textbf{Dimensional Compatibility}: Mating hole and shaft features must have compatible radii. Let $r_q$ and $r_c$ denote the radii of the query and candidate features, respectively. The continuous compatibility weight is
    \begin{equation}
        s_\mathrm{dim}(q,c)=\mathrm{exp}\left(-\frac{(r_q-r_c)^2}{(r_q+r_c)^2}\cdot\frac{1}{\sigma}\right).
    \end{equation}
    This normalized formulation assigns greater support to candidates with closer radii and allows dimensional agreement to re-rank otherwise eligible candidates.

    \item [3.] \textbf{Type-Specific Scoring Logic}: The relevance of dimensional agreement depends on the assembly feature mate type. As described in Section \ref{sec:dataset}, this work includes coaxial hole--hole relations representing connections in which a fastener is absent from the CAD model. The corresponding holes can have different diameters, so a strict radius comparison is not applied to this relation. Let $m_\mathrm{hole}(q,c)$ equal 1 when both features are holes and 0 otherwise. The type-conditioned dimensional potential is
    \begin{equation}
        s_\mathrm{type}(q,c)=m_\mathrm{hole}(q,c)+\left(1-m_\mathrm{hole}(q,c)\right)s_\mathrm{dim}(q,c).
    \end{equation}
    Hole--hole candidates therefore bypass dimensional reweighting, whereas hole--shaft candidates are weighted according to their radii.
\end{itemize}

The engineering compatibility potential and the compatibility-adjusted assembly feature mate score are then defined as
\begin{align}
    \Psi_\mathrm{eng}(q,c) &= m_\mathrm{part}(q,c)s_\mathrm{type}(q,c), \\
    S_\mathrm{mate}(q,c) &= p_\theta(c\mid q,\mathcal{C}_q)\Psi_\mathrm{eng}(q,c).
\end{align}

Here, $p_\theta(c\mid q,\mathcal{C}_q)$ is the learned matching score of candidate $c$ within the candidate set $\mathcal{C}_q$. Because cross-part eligibility and type-conditioned dimensional compatibility are known engineering conditions, their contribution is specified analytically rather than estimated from the training data. The multiplicative fusion admits a qualitative interpretation as posterior reweighting: the learned scores are treated as proportional to prior probabilities over candidates. Let $\mathcal{E}$ denote the engineering compatibility evidence. Under this interpretation, the potential $\Psi_\mathrm{eng}(q,c)$ is treated as proportional to the likelihood of this evidence:
\begin{equation}
    \Psi_\mathrm{eng}(q,c)
    \propto p(\mathcal{E}\mid c,q,\mathcal{C}_q).
\end{equation}
The posterior probabilities under this interpretation are then proportional to
\begin{equation}
    p(c\mid q,\mathcal{C}_q,\mathcal{E})
    \propto p_\theta(c\mid q,\mathcal{C}_q)\Psi_\mathrm{eng}(q,c)
    =S_\mathrm{mate}(q,c).
\end{equation}
This interpretation provides a qualitative probabilistic motivation for the scoring rule. Operationally, $S_\mathrm{mate}(q,c)$ is used as the compatibility-adjusted decision score for training and, during assembly modeling, for candidate ranking and pattern-level score aggregation. No posterior normalization is performed on this score before either use.

The parameter $\sigma$ controls how rapidly dimensional support decreases as the normalized dimensional discrepancy increases. We use $\sigma_\mathrm{train}=1\times10^{-2}$ for training-time candidate scoring and $\sigma_\mathrm{inference}=1\times10^{-5}$ during assembly modeling. The smaller inference value sharpens the dimensional compatibility weight, whereas the larger training value retains graded compatibility over a wider range of sampled candidates.

\subsection{Loss function}
The assembly feature mate prediction module is trained using mean squared error (MSE). For a query feature and its $N$ candidates, the compatibility-adjusted scores from ECF are compared with the one-hot ground-truth vector $\mathbf{y}_\mathrm{gt}$, where $y_{\mathrm{gt},k}=1$ if candidate $k$ is the correct feature and 0 otherwise. The loss is defined as
\begin{equation}
    \mathcal{L}_\mathrm{feat}=\operatorname{MSE}\!\left(
    \bigl(S_\mathrm{mate}(q,c_k)\bigr)_{k=1}^{N},\mathbf{y}_\mathrm{gt}\right).
\end{equation}

This objective trains the attention-based feature mate predictor using scores that incorporate engineering compatibility. ECF contains no learnable parameters, but its compatibility weights affect the gradients propagated to the predictor through the fused scores.

\section{AFP recognition and mate candidate generation}
\label{sec:pattern_prediction}
\label{sec:pattern_problem_def}
The assembly feature mate prediction module produces scored candidates for local feature-to-feature relations. When matching two AFPs, several assembly feature mate candidates may appear plausible when evaluated independently but may not jointly constitute a valid AFP mate. AFP recognition and mate candidate generation are therefore required to evaluate these candidates at the group level. To achieve this, three key issues must be addressed. First, canonical AFPs must be accurately recognized from coplanar feature groups. Second, an AFP mate may admit multiple valid mappings, all of which must be retained as distinct candidates. Third, a small number of non-canonical AFPs fall outside the rectangular and circular layout categories but must still be accommodated in groupwise candidate generation.

As illustrated in Fig.~\ref{fig:framework}, rectangular and circular AFP hypotheses are recognized within each coplanar feature group, and their layout parameters are used to generate the admissible feature mappings. Metric-consistency search handles feature groups that cannot be assigned either layout type. Both paths output AFP mate candidates $\widehat{\mathrm{PM}}$, each comprising a candidate pattern pair and its feature correspondences.

\subsection{Canonical AFP recognition from coplanar feature groups}

Let $G=(\mathcal{F}_G,B_G)$ denote a coplanar feature group, where $\mathcal{F}_G$ is the feature set and $B_G$ is its reference plane. The representative position of each feature is projected onto the local coordinates of $B_G$ as defined in Section~\ref{sec:assembly_feature_pattern}. Canonical recognition produces a set of AFP hypotheses
\begin{equation}
    \mathcal{H}_{\mathrm{can}}(G)=\{P_k \mid \mathcal{F}_{P_k}\subseteq\mathcal{F}_G,\;
    t_k\in\{\mathrm{rect},\mathrm{circ}\}\},
\end{equation}
where $t_k$ denotes the recognized layout family. This operation is not a partition of $G$: features outside the recognized AFPs remain unassigned, and a coplanar group may contain multiple AFPs. Overlap is permitted only between competing recognition hypotheses. In particular, different parameterizations of the same member set, such as the rectangular and circular descriptions of a square four-feature layout, describe one potential AFP rather than two simultaneously accepted instances.

For a rectangular AFP, the projected position $\mathbf{x}_i$ of each constituent feature is modeled as
\begin{equation}
    \mathbf{x}_i=\mathbf{o}+g_i^u s_u\mathbf{u}+g_i^v s_v\mathbf{v}+\mathbf{e}_i,
    \qquad (g_i^u,g_i^v)\in\mathbb{Z}^2,
    \label{eq:rectangular_afp_model}
\end{equation}
where $\mathbf{o}$ is the in-plane origin, $\mathbf{u}$ and $\mathbf{v}$ are orthogonal grid directions, $s_u$ and $s_v$ are the corresponding pitches, and $\mathbf{e}_i$ is the grid-fitting residual. Candidate directions are obtained by clustering the directions of pairwise feature displacements, after which the pitch along each direction is estimated and the projected positions are snapped to integer grid indices. A hypothesis is retained when it contains at least two distinct rows and two distinct columns and its residuals satisfy the geometric tolerance. The occupied grid indices are recorded explicitly, so the representation also accommodates partially populated rectangular layouts.

For a circular AFP, the projected positions are modeled as
\begin{equation}
    \mathbf{x}_i=\mathbf{c}+r
    \begin{bmatrix}
        \cos(\theta_0+k_i\Delta\theta)\\
        \sin(\theta_0+k_i\Delta\theta)
    \end{bmatrix}
    +\mathbf{e}_i,
    \label{eq:circular_afp_model}
\end{equation}
where $\mathbf{c}$ and $r$ are the fitted center and radius, $\theta_0$ is a reference angle, $\Delta\theta$ is the angular pitch, and $k_i$ is an integer angular index. The detector first fits a circle to the projected positions and then verifies both the radial residuals and the equal-angle structure. This second condition prevents an arbitrary set of concyclic points from being classified as a circular AFP.

\subsection{Parametric matching of canonical AFPs}

Once two AFP hypotheses have compatible canonical types, their parameters directly generate possible feature correspondences. Let $P^Q$ denote the smaller query AFP and $P^T$ the target AFP. For rectangular layouts, each admissible mapping is induced by
\begin{equation}
    \mathbf{g}_{\pi(j)}^T=\mathbf{R}\mathbf{g}_j^Q+\mathbf{b},
    \qquad \mathbf{R}\in\mathcal{R}_{\mathrm{grid}},\quad
    \mathbf{b}\in\mathbb{Z}^2,
    \label{eq:rectangular_index_mapping}
\end{equation}
where $\mathbf{g}=(g^u,g^v)^T$, $\mathcal{R}_{\mathrm{grid}}$ contains the eight signed permutation matrices representing the rotations and reflections of a rectangular grid, and $\mathbf{b}$ is an integer grid offset. A transform is admissible only when every transformed query index occurs in the target occupancy set.

For circular layouts with $K$ angular positions, the possible correspondences are generated by
\begin{equation}
    k_{\pi(j)}^T=(\eta k_j^Q+\delta)\bmod K,
    \qquad \eta\in\{-1,1\},\quad \delta\in\{0,\ldots,K-1\},
    \label{eq:circular_index_mapping}
\end{equation}
where $\delta$ represents cyclic phase and $\eta$ distinguishes the two traversal directions. These transformations enumerate the mappings introduced by the intrinsic symmetries of the layout rather than selecting a single arbitrary orientation.

Each generated mapping is subsequently checked against the original feature types and pairwise metric values. All mappings that satisfy the tolerance are retained because a symmetric AFP may admit multiple valid correspondences. If the same correspondence is produced through multiple canonical descriptions, it is represented canonically by its set of feature pairs and duplicate candidates are removed.

For a retained mapping $\pi$, this section outputs the pre-instantiation AFP mate candidate
\begin{equation}
    \widehat{\mathrm{PM}}_{\pi}=(P^T,P^Q,M_{\pi}),
    \qquad
    M_{\pi}=\{(\pi(j),j)\mid j=1,\ldots,|\mathcal{F}_{P^Q}|\}.
    \label{eq:afp_mate_candidate}
\end{equation}
In relation to the AFP mate definition in Section~\ref{sec:assembly_feature_pattern}, $\widehat{\mathrm{PM}}_{\pi}$ contains the pattern pair and its candidate correspondence, whereas the constraint association $\Gamma$ has not yet been constructed.

Rectangular and circular layouts cover the predominant AFP configurations considered in this work. For a feature group that cannot be assigned either canonical type, a supplementary metric-consistency search is applied to generate groupwise correspondence candidates without assigning a canonical layout label. If no valid groupwise correspondence is found, its assembly feature mate candidates remain available for downstream selection.

\section{Assembly constraint generator}

The assembly constraint generator selects assembly feature mate and AFP mate candidates and instantiates the associated assembly constraints for the geometric constraint solver. The selection uses a search procedure similar to the Best-First strategy of Vergez et al. \cite{vergez_multi-part_2025}, with assembly results evaluated by the cumulative scores of the applied assembly feature mates.

\subsection{Unified scoring}
To compare individual assembly feature mate candidates and AFP mate candidates within the same search, we assign each candidate a score $S$. For an individual assembly feature mate $\mathrm{FM}$ connecting features $F_i$ and $F_j$, we average the compatibility-adjusted decision scores obtained with each feature serving as the query:
\begin{equation}
    S_\mathrm{FM}=\frac{S_\mathrm{mate}(i,j)+S_\mathrm{mate}(j,i)}{2}.
\end{equation}
Here, $S_\mathrm{mate}(i,j)$ is the ECF score with $F_i$ as the query and $F_j$ as the candidate, and $S_\mathrm{mate}(j,i)$ is the score with these roles reversed. For an AFP mate ($\mathrm{PM}$) composed of $k$ constituent assembly feature mates $\{\mathrm{FM_1, FM_2, ..., FM_k}\}$, the score is the sum of their averaged scores:
\begin{equation}
    S_\mathrm{PM}=\sum_{i=1}^{k}S_{\mathrm{FM}_i}.
\end{equation}
The score of an AFP mate thus accumulates the matching evidence associated with its layout-consistent feature correspondence.

\subsection{Search procedure}
Candidates are ranked by the scores defined above. The objective is to maximize the sum of the compatibility-adjusted scores of the assembly feature mates applied in the resulting assembly. This sum includes the constituent assembly feature mates of applied AFP mates, with each assembly feature mate counted once.

Each candidate considered for application undergoes an assemblability check. A candidate is accepted and instantiated only if the paired feature types are compatible, the two parts involved do not geometrically interfere, and the candidate does not conflict with the assembly constraints already established in the current assembly state. Accepted candidates are translated into assembly constraints according to their associated geometric entities: cylindrical faces give rise to coaxial constraints, and planar faces give rise to plane-coincidence constraints. These constraints are processed by the geometric constraint solver to obtain the assembly configuration. This interference check covers only the current part pair, whereas $ACC_v$ evaluates all part pairs in the final assembly.

The search first constructs a greedy solution by considering candidates in ranked order and applying those that pass the assemblability check. A tree-based search then explores candidate selections within a Hamming distance of three from this greedy solution. The distance counts candidates whose selection differs from the greedy solution: removing a selected candidate or adding an unselected candidate contributes one. An AFP mate is counted as a single candidate, irrespective of its number of constituent assembly feature mates. The search terminates after this neighborhood has been explored and returns the feasible assembly with the highest cumulative score among the explored solutions.

\section{Results}
\subsection{Dataset}
\label{sec:dataset_results}
Using the automatic annotation pipeline described in Section \ref{sec:dataset}, we constructed a dataset after excluding CAD parts containing more than 500 faces. The resulting dataset contains 529 unique assembly models and 43,646 individual parts, from which the pipeline extracted 106,493 canonical assembly features and 42,618 distinct AFPs. The assembly feature annotations are used to train and evaluate the prediction networks proposed in this work.

The extracted assembly features comprise 93,503 hole-type features, 12,872 shaft-type features, and 118 slot-type features, with holes and shafts together accounting for approximately 99.89\% of the total. The analysis of the Fusion 360 Gallery assembly dataset identifies circular edges and cylindrical faces as the most frequently selected entities for defining joints and attributes this observation to the use of fasteners \cite{willis_joinable_2022}. These observations motivate the focus of the present study on hole and shaft assembly interfaces.

For each prediction network, samples are generated from the dataset and randomly divided into training, validation, and test subsets in an 80\%/10\%/10\% ratio using a fixed random seed. A sample consists of an assembly subgraph for part mate prediction or a query feature with its candidate set for assembly feature mate prediction. Final assembly evaluation and feature-level ablation use a separate set of 16 assemblies excluded from network training, as described below.

\subsection{Evaluation metrics}
To comprehensively evaluate the performance of our framework, we assess both the quality of the final generated assembly and the direct accuracy of our intermediate prediction modules. For the final assembly evaluation, we use the metrics proposed by Vergez et al. \cite{vergez_multi-part_2025}. These metrics are computed on the final constrained assembly $\hat{a}$ generated by our assembly constraint generator and compared against the ground-truth target assembly $a$.

The primary metric is assembly constraint accuracy ($ACC_c$), which measures the fraction of ground-truth assembly constraints $c_j^a$ from the target assembly $a$ that are present in the set of equivalent assembly constraints $C_{\hat{a}}$ found in our predicted assembly $\hat{a}$. Formally:

\begin{equation}
    ACC_{c} = \frac{1}{N_a} \sum_{a=1}^{N_a} \left( \frac{1}{N_c^a} \sum_{j=1}^{N_c^a} \mathbb{I} [c_j^a \in C_{\hat{a}}] \right)
\end{equation}

where $N_a$ is the total number of assemblies, $N_c^a$ is the number of assembly constraints in a target assembly $a$, and $\mathbb{I}$ is the indicator function. Two assembly constraints are considered equivalent if they can form the same kinematic pairs.

Assembly constraint loop accuracy ($ACC_{\ell}$) measures the fraction of assembly constraint cycles in the target assembly that are recovered in the predicted assembly, averaged over the evaluated assemblies:
\begin{equation}
            ACC_{\ell} = \frac{1}{N_a} \sum_{a=1}^{N_a} \left( \frac{1}{N_{\ell}^a} \sum_{j=1}^{N_{\ell}^a} \mathbb{I} \left[ \ell_j^a \in \mathcal{L}_{\hat{a}} \right] \right)
\end{equation}
where $N_{\ell}^a$ is the number of cycles in the target assembly $a$, $\ell_j^a$ denotes its $j$th cycle, and $\mathcal{L}_{\hat{a}}$ is the set of cycles in the predicted assembly $\hat{a}$.

Finally, $ACC_v$ is the fraction of non-intersecting part pairs, averaged over the evaluated assemblies:
\begin{equation}
    ACC_v = 1 - \frac{1}{N_a} \sum_{a=1}^{N_a} \left( \frac{\sum_{i=1}^{N_p^a - 1} \sum_{j=i+1}^{N_p^a} \mathbb{I} \left[ P_i \cap P_j \right]}{N_p^a (N_p^a - 1) / 2} \right)
\end{equation}
where $N_p^a$ is the number of parts in the assembly $a$.

In addition to these end-to-end metrics, we evaluate the predictive accuracy of the pretraining and feature-level tasks. For the pretraining task of part mate prediction, accuracy is evaluated as a binary classification problem. The continuous output scores are thresholded at 0.5; scores greater than 0.5 are classified as a positive part mate, and accuracy is computed against the ground-truth part mate matrix. For the assembly feature mate prediction module, accuracy is measured as a top-1 classification task. For a given query $q$ with $N$ candidates, let $\mathbf{s}_q \in \mathbb{R}^N$ be the vector of output scores, and let $i_{gt}$ be the index of the ground-truth feature. We define the set of predicted top indices as $I_{max} = \{ i \mid \mathbf{s}_{q,i} = \max(\mathbf{s}_q) \}$. The accuracy for this query, $acc_q$, is defined as:
\begin{equation}
    acc_q = \frac{\mathbb{I}[i_{gt} \in I_{max}]}{|I_{max}|}
\end{equation}
This formulation correctly handles ties: if the ground-truth is one of $N$ identical maximum scores, the accuracy is $1/N$. The final $ACC_{f}$ is the average of $acc_q$ over all queries in the test set.

\subsection{Experimental results}

The proposed framework was implemented using PyTorch on a server equipped with an Intel Core i9-13900K CPU and an NVIDIA RTX 4090 GPU. SolidWorks was utilized via API for dataset construction, geometric solving, and visualization. The part encoder was first pretrained through part mate prediction and then frozen during assembly feature mate prediction training. The networks were trained for 500 epochs using the Adam optimizer with an initial learning rate of $1\times10^{-3}$.

We compare our method with JoinABLe \cite{willis_joinable_2022} and the framework of Vergez et al. \cite{vergez_multi-part_2025}, using the baseline results reported in the latter study. The modified assemblies used in that study are not publicly available. We therefore evaluate our method on six corresponding original models from the Autodesk Fusion 360 Gallery, with geometric modeling errors in the original data corrected. These six assemblies constitute the \textit{assembly comparison set}.

To compare with Vergez et al. under conditions where assembly interface information is available, we provide our method with face-level labels indicating which faces participate in assembly. These labels are derived from the ground-truth assembled models. Assembly feature extraction remains unchanged, and the extracted features are subsequently filtered according to the labels of their constituent faces. This setting is denoted as \textit{Ours (gt face input)}. The complete automatic workflow, denoted as \textit{Ours}, operates directly on the input B-Rep parts without these labels.

All three variants of our method in Table~\ref{tab:comparison} use a post-processing search procedure similar to the Best-First strategy of Vergez et al. \cite{vergez_multi-part_2025}, with the same Hamming distance limit of three across variants. The objective is to maximize the sum of the compatibility-adjusted scores of the assembly feature mates applied in the resulting assembly. AFP mate candidates contribute the scores of their constituent assembly feature mates, with each assembly feature mate counted once. Part mate scores are used only to break ties in candidate ranking and do not contribute to this sum.

\begin{figure}[htbp]
    \centering
    \includegraphics[width=1\linewidth]{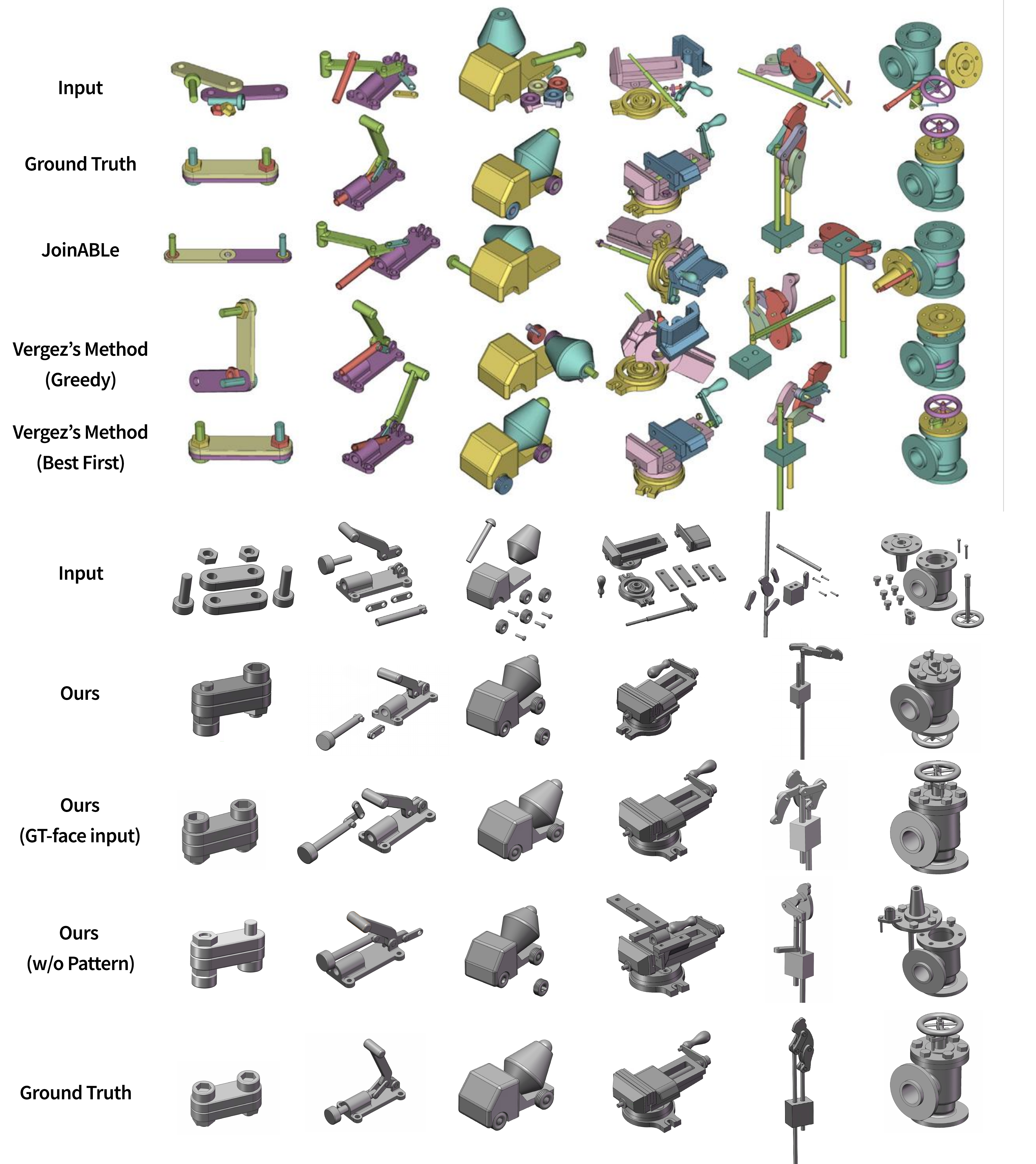}
    \caption{Qualitative comparison on the assembly comparison set. The top section (colored models) shows the baseline results for JoinABLe \cite{willis_joinable_2022} and Vergez et al. \cite{vergez_multi-part_2025}, as reported in \cite{vergez_multi-part_2025}. The bottom section (gray models) shows the results of our method on the six corresponding original assemblies.}
    \label{fig:compare_result}
\end{figure}

Table~\ref{tab:comparison} shows that, with face labels, our method achieves $ACC_c=81.78\%$, $ACC_{\ell}=65.38\%$, and $ACC_v=100\%$, compared with the reported values of $72.94\%$, $32.86\%$, and $81.81\%$ for the Best-First strategy of Vergez et al. Without face labels, the complete workflow achieves $ACC_c=74.79\%$, $ACC_{\ell}=34.61\%$, and $ACC_v=92.86\%$. By combining automatic assembly feature extraction, assembly feature mate prediction with part context and ECF, and AFP recognition and mate candidate generation, the framework thus attains an $ACC_c$ comparable to the reported Best-First result while operating without prior face labels. Figure~\ref{fig:compare_result} presents the corresponding assembly results.

These results support the approach motivated in Section~\ref{sec:introduction}: changing the prediction targets from individual assembly constraints to assembly feature mates, so that complementary constraints are considered together as a local assembly relation. The effect of this choice on the candidate space is analyzed in Section~\ref{sec:candidate_space_reduction}.

\begin{table}[htbp]
\centering
\caption{Quantitative comparison of automatic assembly approaches.}
\label{tab:comparison}
\renewcommand{\arraystretch}{1.2} %
\setlength{\tabcolsep}{5mm}{
    \begin{tabular}{lccc}
    \toprule
    Method & $ACC_c \uparrow$ & $ACC_{\ell} \uparrow$ & $ACC_v \uparrow$ \\
    \midrule
    JoinABLe \cite{willis_joinable_2022} & 24.00\% & 0.00\% & 61.29\% \\
    Vergez et al.'s greedy \cite{vergez_multi-part_2025} & 16.15\% & 0.00\% & 70.99\% \\
    Vergez et al.'s Best-First \cite{vergez_multi-part_2025} & {72.94\%} & 32.86\% & 81.81\% \\
    \midrule %
    Ours w/o pattern & 38.93\% & 7.69\% & 83.93\% \\
    Ours (gt face input) & \textbf{81.78}\% & \textbf{65.38}\% & \textbf{100}\% \\
    Ours & 74.79\% & {34.61\%} & {92.86\%} \\
    \bottomrule
    \end{tabular}
}
\end{table}

\begin{figure}[htbp]
    \centering
    \includegraphics[width=1\linewidth]{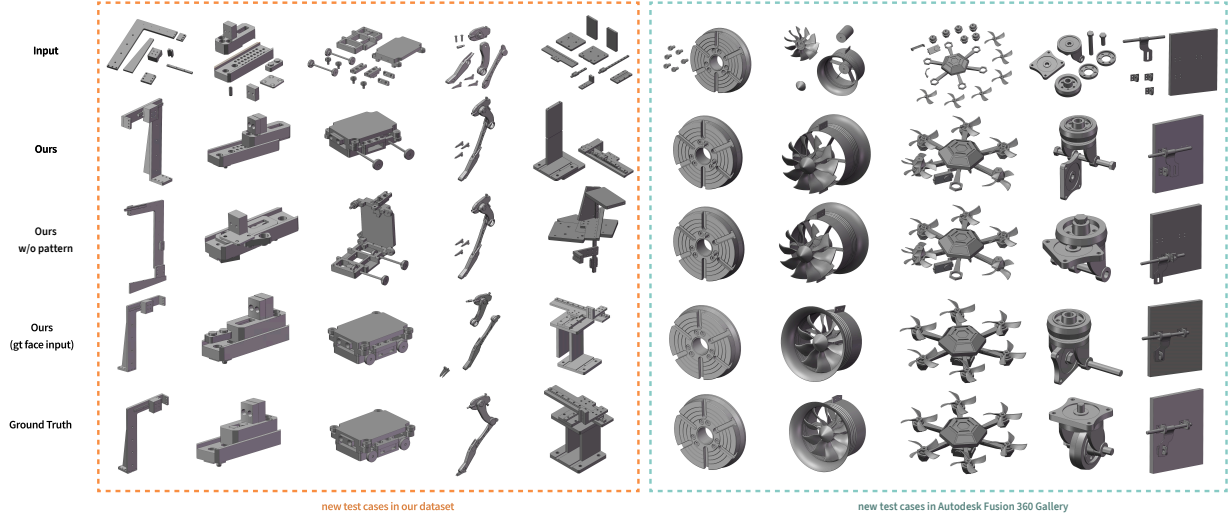}
    \caption{Qualitative results on the ten additional assemblies in the extended test set.
    }
    \label{fig:new_results}
\end{figure}

To further evaluate our method, the evaluation is extended to ten additional assemblies without fine-tuning. Together with the assembly comparison set, they form the \textit{extended test set} of 16 assemblies: five from our dataset and eleven from the Autodesk Fusion 360 Gallery. All 16 assemblies were excluded from network training and are used for both assembly-level evaluation and feature-level ablation. As reported in Table~\ref{tab:extended benchmark}, the complete workflow achieves $ACC_c=64.09\%$, $ACC_{\ell}=32.14\%$, and $ACC_v=82.27\%$ on this set. With face labels, the corresponding values are $80.93\%$, $65.48\%$, and $100\%$. Figure~\ref{fig:new_results} shows the ten additional assemblies.

\begin{table}[htbp]
    \centering
    \caption{Quantitative results on the extended test set of 16 assemblies.}
    \begin{tabular}{lccc}
    \toprule
    Method & $ACC_c $ & $ACC_{\ell}$ & $ACC_v $ \\
    \midrule
    Ours w/o pattern & 31.77\% & 9.68\% & 56.74\% \\
    Ours (gt face input) & 80.93\% & 65.48\% & 100\% \\
    Ours & 64.09\% & 32.14\% & 82.27\% \\
    \bottomrule
    \end{tabular}
    \label{tab:extended benchmark}
\end{table}

The contribution of AFP recognition and mate candidate generation is evaluated by removing this module while retaining assembly feature mate prediction and the same post-processing search settings. On the assembly comparison set, removing the module reduces $ACC_c$ from $74.79\%$ to $38.93\%$; on the extended test set, it reduces $ACC_c$ from $64.09\%$ to $31.77\%$. Assembly constraint loop accuracy also decreases, from $34.61\%$ to $7.69\%$ and from $32.14\%$ to $9.68\%$, respectively. These results show the value of using AFP layout information to constrain correspondences between constituent features. Assembly features can have similar or identical local geometry, whereas their relative arrangement provides additional information about how they should match. Requiring correspondences to satisfy the AFP layout reduces incorrect matches between such features and improves the resulting assembly accuracy. The qualitative comparisons in Figs.~\ref{fig:compare_result} and~\ref{fig:new_results} illustrate the resulting improvements in automatic assembly modeling.

A feature-level ablation evaluates the contribution of ECF on the 16 assemblies in the extended test set (Table~\ref{tab:ablation}). The assembly feature mate prediction module is retrained without ECF and evaluated with ECF removed. The full model achieves $ACC_f=76.27\%$, compared with $55.79\%$ for the retrained model without ECF, a difference of 20.48 percentage points. This result supports the contribution of explicit engineering compatibility to assembly feature mate prediction on these assemblies.

\begin{table}[htbp]
\centering
\caption{Assembly feature mate prediction accuracy ($ACC_{f}$) on the extended test set of 16 assemblies. The model without ECF is retrained and evaluated with ECF removed.}
\label{tab:ablation}
\begin{tabular}{l c c}
\toprule
\textbf{Evaluation Setting} & \textbf{Assembly Feature Mate Accuracy ($ACC_{f}$)}& \textbf{Drop (points)} \\
\midrule
\textbf{Ours (Full Model)} & 76.27\% & - \\
w/o Engineering Compatibility Fusion (ECF) & 55.79\%& 20.48\\
\bottomrule
\end{tabular}
\end{table}

The benefit of ground-truth face labels is related to ambiguity in selecting the features that participate in the target assembly. Among several geometrically similar features that provide plausible matches, only one may actually participate in the ground-truth assembly. In such cases, filtering extracted features according to the face labels excludes the nonparticipating features, directly resolving the ambiguity arising from their competing matches. This helps explain the increase in $ACC_c$ from $64.09\%$ to $80.93\%$ on the extended test set when ground-truth face labels are supplied.

In practical assembly modeling, engineers need to assess whether an automatically generated assembly is correct as a whole. Assembly constraint accuracy measures the recovery of individual constraints, but does not directly indicate how many assemblies are completely correct or how many parts are involved in the remaining corrections. To address these questions, Table~\ref{tab:assembly_correction} reports the proportions of completely correct assemblies and assemblies requiring correction of the relations involving at most two parts across the 16 assemblies in the extended test set. These outcomes are determined by manually inspecting the solved assemblies, including their assembly constraints, kinematic pairs, and part configurations. An assembly is considered completely correct only if its parts are correctly installed, all target assembly constraints are recovered with no additional constraints, and the intended motion is preserved. Correcting a part means revising its assembly relations to recover the target assembly, and the latter proportion includes completely correct assemblies.

Without face labels, no assembly is completely correct, but five ($31.25\%$) require correction of the relations involving at most two parts. With face labels, nine assemblies ($56.25\%$) are completely correct, and two more require correction of the relations involving at most two parts, bringing the proportion with at most two parts requiring correction to $68.75\%$. These results show that, in some assemblies, the required corrections involve only a small number of parts. Assessing the practical utility of automatic assembly modeling therefore requires considering how errors are distributed across assemblies and their parts, in addition to average assembly constraint accuracy. The assembly situations that remain challenging for the current method and the associated limitations are discussed further below.

\begin{table}[htbp]
    \centering
    \caption{Assembly-level results based on manual inspection of the extended test set. Correcting a part means revising its assembly relations. The second row includes completely correct assemblies.}
    \label{tab:assembly_correction}
    \begin{tabular}{p{0.43\linewidth}cc}
        \toprule
        Assembly outcome & Ours & \makecell{Ours\\(gt face input)} \\
        \midrule
        Completely correct & 0\% (0/16) & 56.25\% (9/16) \\
        Relations of at most two parts require correction & 31.25\% (5/16) & 68.75\% (11/16) \\
        \bottomrule
    \end{tabular}
\end{table}

\FloatBarrier
\section{Discussion and limitations}
\subsection{Effect of assembly feature representation on candidate space}
\label{sec:candidate_space_reduction}

As discussed in the Introduction, treating assembly constraints as individual prediction targets leaves the associations between complementary constraints unspecified. Without these associations, candidate plane-coincidence and coaxial constraints can be freely combined, giving rise to a large candidate space. To quantify the effect of local interface associations, we first examine the reduction in planar--cylindrical face combinations within individual parts and then consider its effect on cross-part candidate formation.

For part $i$, let $n_{\mathrm{pl}}^i$, $n_{\mathrm{cyl}}^i$, and $n_{\mathrm{AF}}^i$ denote the numbers of planar faces, cylindrical faces, and extracted hole and shaft assembly features, respectively. The unrestricted planar--cylindrical face combinations number $n_{\mathrm{pl}}^i n_{\mathrm{cyl}}^i$, whereas each extracted assembly feature associates a cylindrical face with a reference planar face. Face counts are obtained from the B-Rep geometry without using existing mate definitions, and feature counts are obtained from the assembly feature extraction module. For a nonzero reference count, the reduction on part $i$ is
\begin{equation}
    1-\frac{n_{\mathrm{AF}}^i}{n_{\mathrm{pl}}^i n_{\mathrm{cyl}}^i}.
\end{equation}
Across the 139 parts in the 16 evaluated assemblies, the mean per-part reduction is $76.55\%$, with each part weighted equally.

For a candidate part pair $(i,j)$, all unfiltered pairings of the extracted features yield $n_{\mathrm{AF}}^i n_{\mathrm{AF}}^j$ assembly feature mate candidates. In the reference representation, $n_{\mathrm{pl}}^i n_{\mathrm{pl}}^j$ candidate plane-coincidence constraints can be freely combined with $n_{\mathrm{cyl}}^i n_{\mathrm{cyl}}^j$ candidate coaxial constraints. This reference counts possible combinations without assuming that an existing method explicitly enumerates them. For a nonzero reference count, the ratio is
\begin{equation}
    \frac{n_{\mathrm{AF}}^i n_{\mathrm{AF}}^j}
    {(n_{\mathrm{pl}}^i n_{\mathrm{pl}}^j)(n_{\mathrm{cyl}}^i n_{\mathrm{cyl}}^j)}
    =
    \underbrace{
        \frac{n_{\mathrm{AF}}^i}{n_{\mathrm{pl}}^i n_{\mathrm{cyl}}^i}
    }_{\text{Ratio for part }i}
    \;\cdot\;
    \underbrace{
        \frac{n_{\mathrm{AF}}^j}{n_{\mathrm{pl}}^j n_{\mathrm{cyl}}^j}
    }_{\text{Ratio for part }j}.
\end{equation}
Each factor is the retained fraction of planar--cylindrical face combinations on one part. Their product determines the retained fraction for the part pair, so local restrictions on both parts jointly reduce the cross-part candidate space. This relation applies to the two parts in each pair; the mean per-part reduction alone does not determine the aggregate reduction across part pairs.

\FloatBarrier

The combinatorial growth of the candidate space is a central difficulty in automatic assembly modeling. Previous studies address this challenge under different task settings and with different sources of assembly information. AutoMate \cite{jones_automate_2021} and JoinABLe \cite{willis_joinable_2022} focus on predicting assembly relations between a given pair of parts. Vergez et al. \cite{vergez_multi-part_2025} extend prediction to multi-part assemblies, using predefined assembly faces. In our framework, we change the prediction targets from individual assembly constraints to assembly feature mates, using local interface associations extracted directly from the input B-Rep geometry to restrict candidate formation. The above analysis quantifies this restriction within individual parts and explains how it carries over to cross-part candidates. This change in representation supports automatic assembly modeling without requiring predefined assembly faces on the input parts. Complete assembly construction still requires selecting mutually compatible relations and determining a consistent configuration; the candidate counts do not measure feasible complete assembly states or a proportional reduction in runtime.

\subsection{Multi-solution ambiguity}
\label{sec:multi_solution_ambiguity}

\begin{figure}[htbp]
    \centering
    \includegraphics[width=\linewidth]{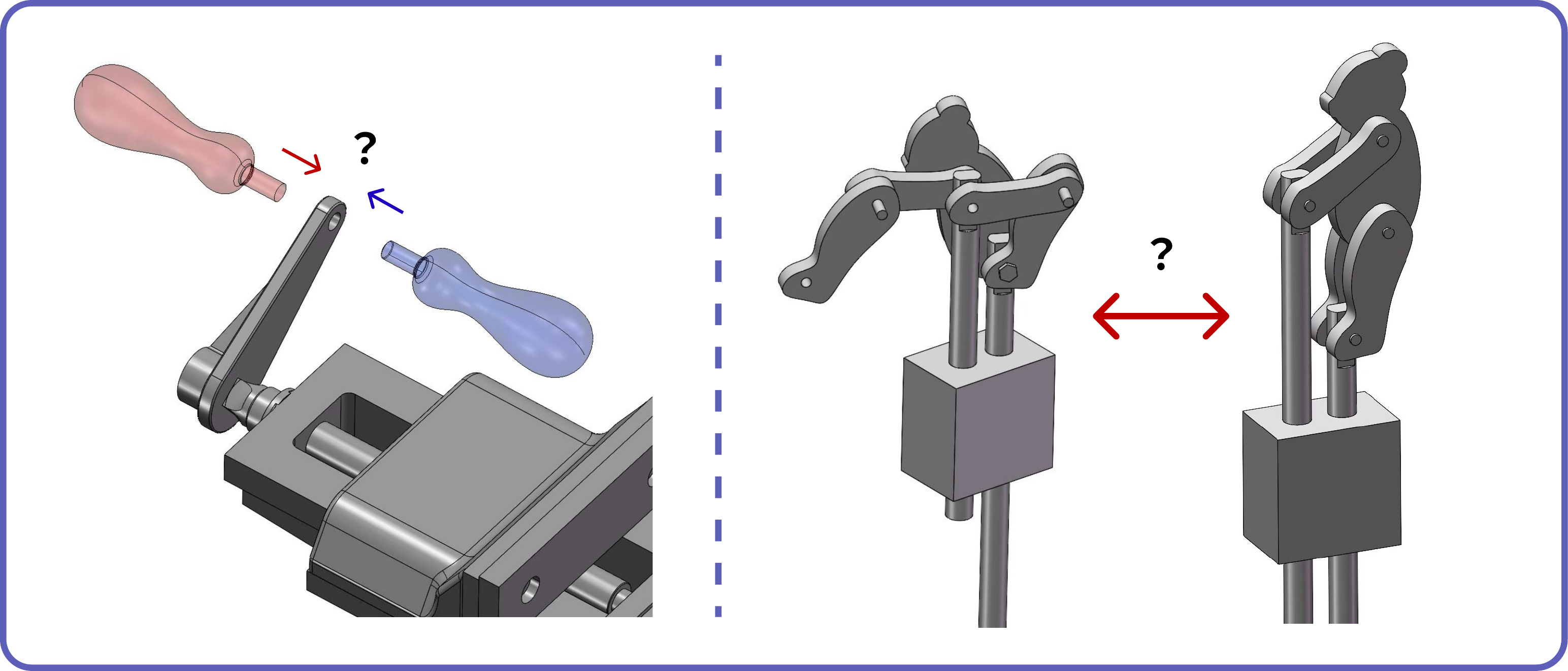}
    \caption{Examples of multi-solution ambiguity. Left: the two colored handles illustrate alternative mounting directions for the same handle, which can be inserted into the through hole from either side. Right: similar hole and shaft interfaces permit different ways of connecting the links in a multi-link assembly. In each example, only one configuration corresponds to the ground-truth assembly: the mounting direction shown by the red handle on the left and the rightmost toy bear configuration on the right.}
    \label{fig:multi_solution_ambiguity}
\end{figure}

Reducing the candidate space through assembly feature representation still leaves ambiguity in selecting the relations that recover the target assembly. Assemblies often contain local features with similar or identical shapes and dimensions, giving rise to multiple assembly feature mate candidates that satisfy local mating conditions. For example, a handle may be inserted into a through hole from either side (Fig.~\ref{fig:multi_solution_ambiguity}, left), while holes and shafts with the same dimensions in a multi-link assembly may permit different ways of connecting the links (Fig.~\ref{fig:multi_solution_ambiguity}, right). Such candidates can therefore lead to different mounting directions or connection structures, only some of which correspond to the target assembly. Local geometric compatibility alone is insufficient to determine the intended matches in these cases.

The framework uses parent-part context and AFP layout information to help distinguish competing matches, but ambiguity remains in some assembly situations. Symmetry provides a typical example. Although AFP layouts restrict correspondences between similar features, a symmetric layout may admit multiple feature correspondences that satisfy the same layout conditions. For individual assembly feature mate candidates, symmetry in the parent parts can also limit the contextual information available to distinguish locally similar features. The current model does not consistently distinguish the intended matches in such cases. Thus, despite the use of part context and AFP layouts, resolving matching ambiguity remains a limitation of the framework: geometrically plausible candidates may still be selected in place of the relations required by the target assembly.

\FloatBarrier

\subsection{Limitations of the assembly feature representation}
\label{sec:feature_representation_limitations}

The proposed representation uses local interface associations to group complementary constraints into assembly feature mates, restricting candidate formation as analyzed in Section~\ref{sec:candidate_space_reduction}. Although this organization supports automatic assembly modeling, the current formulation does not cover interfaces formed by several parts or individual constraints that do not belong to a complementary group.

\begin{figure}[htbp]
    \centering
    \includegraphics[width=\linewidth]{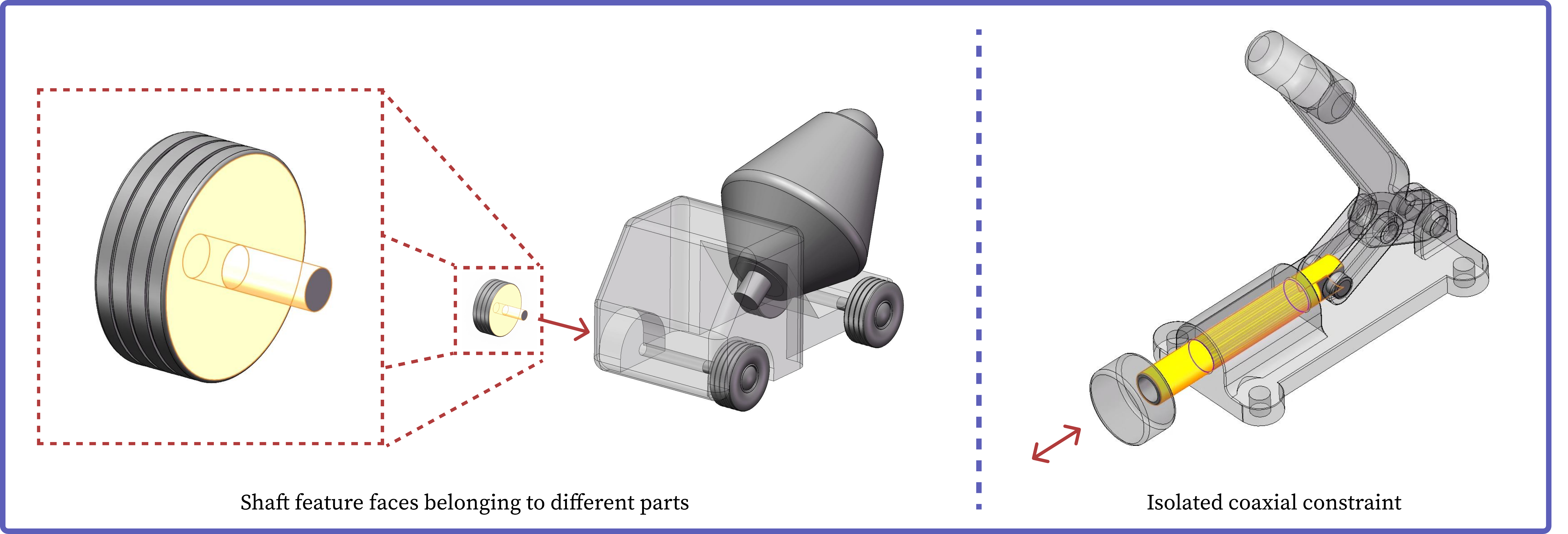}
    \caption{Examples illustrating limitations of the assembly feature representation. Left: the highlighted cylindrical face and planar face belong to different parts and jointly form a shaft feature after assembly. Right: the highlighted cylindrical faces participate in an isolated coaxial constraint; the double-headed arrow indicates relative motion along the common axis.}
    \label{fig:feature_representation_limitations}
\end{figure}

In some assemblies, the faces that define an assembly feature belong to different parts. Figure~\ref{fig:feature_representation_limitations} (left) illustrates a shaft feature whose cylindrical face and planar face are provided by separate parts. This feature emerges at an intermediate stage of assembly, when the parts establish the relative configuration in which these faces jointly form the interface. It therefore does not exist as a complete feature on either initial part and cannot be extracted from either part in advance. The current framework extracts features from the individual input parts before assembly and thus does not capture these interfaces as they form. Their identification depends on the evolving assembly state, extending beyond feature extraction from the initial part geometry. This observation suggests that assembly feature recognition and assembly modeling may be interdependent.

Some faces participate in assembly constraints without forming the grouped interfaces used by the current representation. Figure~\ref{fig:feature_representation_limitations} (right) shows an isolated coaxial constraint between cylindrical faces, with relative motion along the common axis indicated by the arrow. This local relation is expressed by a single assembly constraint, whereas the current assembly feature mate formulation associates mating features with a group of complementary constraints. Such individual relations are therefore not directly covered by the current prediction targets. Together, these examples show that the representation is limited both by its reliance on interfaces extracted from individual parts and by its use of grouped constraints to express local assembly relations.

\FloatBarrier

\subsection{Influence of initial part poses on constraint solving}
\label{sec:initial_part_poses}

A correct assembly configuration requires the parts to be free of geometric interference. During part installation, the framework checks each candidate for conflicts with the existing assembly constraints and for geometric interference between its two parts in the solved configuration. Our experiments use the default initial part poses assigned by SolidWorks when loading the parts. For assemblies with moving parts, these initial poses can influence the configuration obtained by the geometric constraint solver. Even when a candidate specifies the correct assembly relation, the solver may place a moving part in a position where it interferes with the other part in the candidate relation. The interference check may then cause the system to reject this candidate, although a configuration without interference exists for the same relation. Thus, interference in the configuration reached from a given initialization can be mistaken for an invalid assembly relation, causing a correct candidate to be discarded and limiting the recovery of the target assembly.

\section{Conclusion}

This work presents a data-driven framework that uses assembly feature mates as prediction targets and an attention-based network to predict them, thereby making it feasible to perform automatic assembly modeling directly from unassembled B-Rep parts without predefined assembly faces. Incorporating AFPs further improves the accuracy of automatic assembly modeling. We also constructed a dataset of 529 assemblies through an automatic annotation pipeline, supplying the assembly feature annotations required for network training.

The complete workflow achieves assembly constraint accuracies of $74.79\%$ on the assembly comparison set and $64.09\%$ on the extended test set of 16 assemblies. With ground-truth face labels supplied as additional input, the corresponding accuracies increase to $81.78\%$ and $80.93\%$, respectively. For comparison, Vergez et al.\ \cite{vergez_multi-part_2025} reported an assembly constraint accuracy of $72.94\%$ for their Best-First strategy on modified versions of the assemblies in the comparison set. Ablation results support the role of AFP layout information in improving automatic assembly modeling accuracy and that of ECF in improving assembly feature mate prediction accuracy. The candidate-space analysis explains how local interface associations restrict candidate formation, supporting the choice of assembly feature mates as prediction targets.

Although multi-solution ambiguity and the limited coverage of the current representation remain challenges, this work shows how local interface associations and AFP layout information can support automatic assembly modeling directly from unassembled B-Rep parts.

\section*{CRediT authorship contribution statement}
\textbf{Ziheng Dai:} Writing – original draft, Conceptualization, Visualization, Validation, Software, Resources, Methodology, Investigation. \textbf{Wanbin Pan:} Writing - Review \& Editing, Methodology, Validation, Supervision. \textbf{Zhihong Tang:} Conceptualization, Methodology, Writing - Review \& Editing. \textbf{Shuming Gao:} Conceptualization, Methodology, Resources, Investigation, Writing - Review \& Editing, Supervision, Project administration.

\section*{Declaration of competing interest}
The authors declare that they have no known competing financial interests or personal relationships that could have appeared to influence the work reported in this paper.

\section*{Acknowledgment}
This work is financially supported by the National Natural Science Foundation of China (No. 62472134).

\newpage

 \bibliographystyle{elsarticle-num}
 \bibliography{reference}

\end{document}